%% file: main.tex
\def\year{2021}\relax
\documentclass[letterpaper]{article} 
\usepackage[dvipsnames]{xcolor}
\usepackage{aaai21}  
\usepackage{times}  
\usepackage{helvet} 
\usepackage{courier}  
\usepackage[hyphens]{url}  
\usepackage{graphicx} 
\usepackage{natbib}  
\usepackage{caption} 
\usepackage{subcaption}
\newlength\lengtha 
\newlength\lengthb 
\newlength\lengthc 
\newlength\lengthd 

\usepackage{url}
\input{settings.tex}

\input{math_commands.tex}

\newcommand\ours{\textsc{AttAttr}}
\title{Self-Attention Attribution: \\ Interpreting Information Interactions Inside Transformer}
\author {
    Yaru Hao,\textsuperscript{\rm 12}\thanks{\ \  Contribution during internship at Microsoft Research.}
    Li Dong,\textsuperscript{\rm 2}
    Furu Wei,\textsuperscript{\rm 2}
    Ke Xu \textsuperscript{\rm 1} \\
}
\affiliations {
    \textsuperscript{\rm 1} Beihang University \\
    \textsuperscript{\rm 2} Microsoft Research \\
    \texttt{\{haoyaru@,kexu@nlsde.\}buaa.edu.cn} \\ \texttt{\{lidong1,fuwei\}@microsoft.com}
}

\begin{document}

\maketitle

\begin{abstract}
The great success of Transformer-based models benefits from the powerful multi-head self-attention mechanism, which learns token dependencies and encodes contextual information from the input. Prior work strives to attribute model decisions to individual input features with different saliency measures, but they fail to explain how these input features interact with each other to reach predictions. In this paper, we propose a self-attention attribution method to interpret the information interactions inside Transformer. We take BERT as an example to conduct extensive studies. Firstly, we apply self-attention attribution to identify the important attention heads, while others can be pruned with marginal performance degradation. Furthermore, we extract the most salient dependencies in each layer to construct an attribution tree, which reveals the hierarchical interactions inside Transformer. Finally, we show that the attribution results can be used as adversarial patterns to implement non-targeted attacks towards BERT.
\end{abstract}

\section{Introduction}
\label{sec:intro}

Transformer~\cite{transformer} is one of state-of-the-art NLP architectures.
For example, most pre-trained language models~\cite{bert,roberta,unilm,unilmv2,electra,xlmr,xnlg,infoxlm} choose stacked Transformer as the backbone network.
Their great success stimulates broad research on interpreting the internal black-box behaviors.
Some prior efforts aim at analyzing the self-attention weights generated by Transformer~\cite{what-bert-look,revealing-dark}. In contrast, some other work argues that self-attention distributions are not directly interpretable~\cite{is-attention-interpret,att-not, Brunner2020On}. 
Another line of work strives to attribute model decisions back to input tokens~\cite{ig,deeplift,lrp}. 
However, most previous attribution methods fail on revealing the information interactions between the input words and the compositional structures learnt by the network.

To address the above issues, we propose a self-attention attribution method (\ours{}) based on integrated gradient~\cite{ig}.
We conduct experiments for BERT~\cite{bert} because it is one of the most representative Transformer-based models.
Notice that our method is general enough, and can be applied to other Transformer networks without significant modifications.
Results show that our method well indicates the information flow inside Transformer, which makes the self-attention mechanism more interpretable.

Firstly, we identify the most important attention connections in each layer using \ours{}.
We find that attention weights do not always correlate well with their contributions to the model prediction.
We then introduce a heuristic algorithm to construct self-attention attribution trees, which discovers the information flow inside Transformer.
In addition, a quantitative analysis is applied to justify how much the edges of an attribution tree contribute to the final prediction.

Next, we use \ours{} to identify the most important attention heads and perform head pruning.
The derived algorithm achieves competitive performance compared with the Taylor expansion method~\cite{are16headbetterthan1}.
Moreover, we find that the important heads of BERT are roughly consistent across different datasets as long as the tasks are homogeneous.

Finally, we extract the interaction patterns that contribute most to the model decision, and use them as adversarial triggers to attack BERT-based models.
We find that the fine-tuned models tend to over-emphasize some word patterns to make the prediction, which renders the prediction process less robust.
For example, on the MNLI dataset, adding one adversarial pattern into the premise can drop the accuracy of \texttt{entailment} from $82.87\%$ to $0.8\%$.
The results show that \ours{} not only can interpret the model decisions, but also can be used to find anomalous patterns from data.

The contributions of our work are as follows:
\begin{itemize}[leftmargin=*]
\setlength\itemsep{0em}
\item We propose to use self-attention attribution to interpret the information interactions inside Transformer.
\item We conduct extensive studies for BERT. We present how to derive interaction trees based on attribution scores, which visualizes the compositional structures learnt by Transformer.
\item We show that the proposed attribution method can be used to prune self-attention heads, and construct adversarial triggers.
\end{itemize}

\begin{figure}[t!]
\centering
\begin{subfigure}[t]{0.48\linewidth}
\centering
\includegraphics[width=\textwidth]{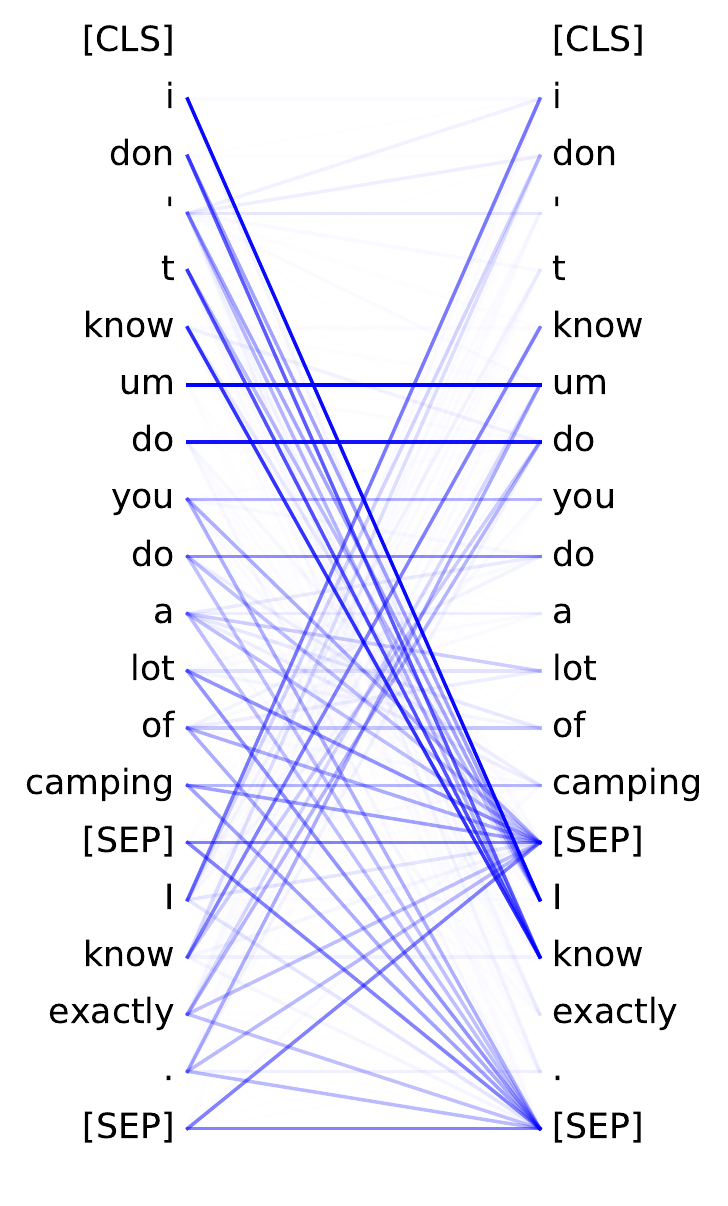}
\caption{Attention Score}
\label{fig:y equals x}
\end{subfigure}
\hfill
\begin{subfigure}[t]{0.48\linewidth}
\centering
\includegraphics[width=\textwidth]{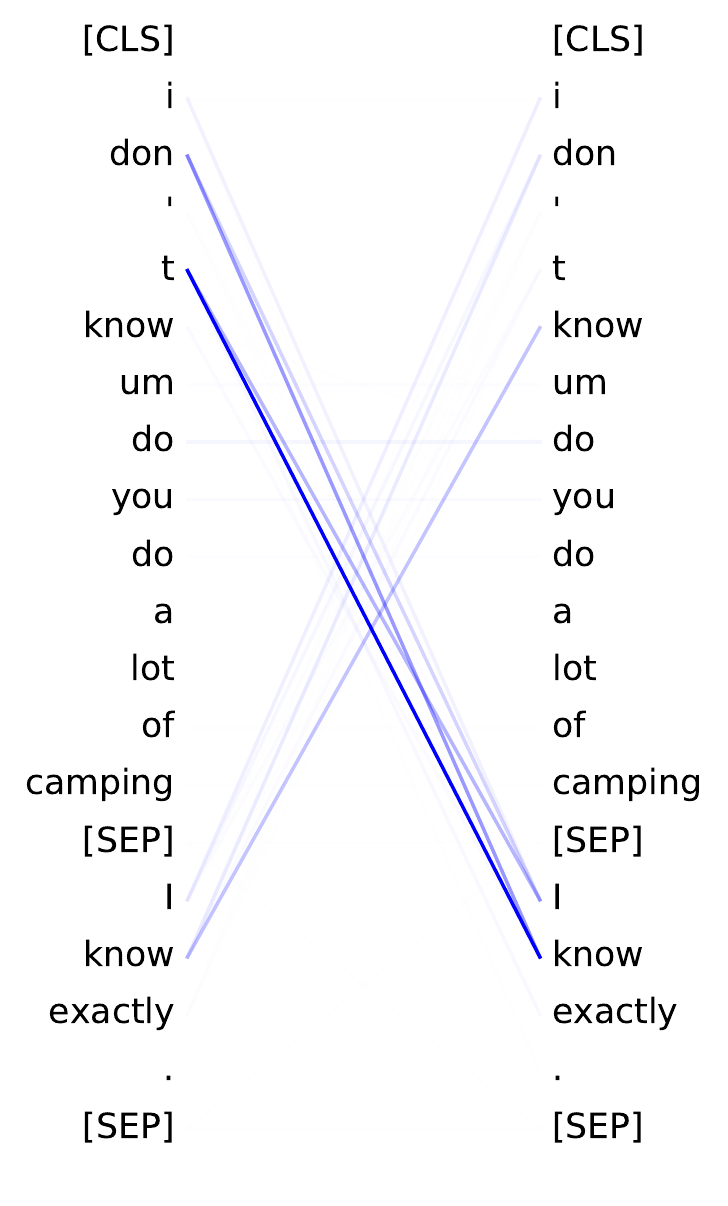}
\caption{Attribution Score}
\label{fig:three sin x}
\end{subfigure}
\caption{Attention score (left) and attribution score (right) of a single head in BERT. The color is darker for larger values.
The model prediction for the sentence from MNLI dataset is \texttt{contradiction}.
\ours{} tends to identify more sparse word interactions that contribute to the final model decision.
}
\label{fig:attig_map} 
\end{figure}

\section{Background}

\subsection{Transformer}
\label{sec:transformer}

Transformer~\cite{transformer} is a model architecture relying on the attention mechanism.
Given input tokens $\{{x}_i\}_{i=1}^{|x|}$, we pack their word embeddings to a matrix ${X}^0 = [{x}_1, \cdots, {x}_{|x|}]$.
The stacked $L$-layer Transformer computes the final output via ${X}^l = \mathrm{Transformer}_{l}({X}^{l-1}), l \in [1,L]$.

The core component of a Transformer block is multi-head self-attention.
The $h$-th self-attention head is described as:
\begin{align}
{Q}_{h} &= X {W}_{h}^Q ,~ {K} = X {W}_{h}^K ,~ {V} = X {W}_{h}^V \\
&A_{h} = \softmax (\frac{Q_{h} K_{h}^{\intercal}}{\sqrt{d_k}}) \label{eq:att:score} \\
&H_{h} = \mathrm{AttentionHead}(X) = A_{h} V_{h} \label{eq:att:head}
\end{align}
where $Q,K \in {\mathbb{R}}^{n\times d_k}$, $V\in {\mathbb{R}}^{n\times d_v}$, and the score $A_{i,j}$ indicates how much attention token $x_i$ puts on $x_j$.
There are usually multiple attention heads in a Transformer block. The attention heads follow the same computation despite using different parameters.
Let $|h|$ denote the number of attention heads in each layer, the output of multi-head attention is given by $\mathrm{MultiH}(X) = [ H_1, \cdots, H_{|h|} ] W^{o}$, where $W^{o} \in {\mathbb{R}}^{|h|d_v\times d_x}$, $[\cdot]$ means concatenation, and $H_i$ is computed as in \eqform{eq:att:head}.

\subsection{BERT}
\label{sec:bert}

We conduct all experiments on BERT~\cite{bert}, which is one of the most successful applications of Transformer.
The pretrained language model is based on bidirectional Transformer, which can be fine-tuned towards downstream tasks.
Notice that our method can also be applied to other multi-layer Transformer models with few modifications. 
For single input, a special token \sptk{CLS} is added to the beginning of the sentence, and another token \sptk{SEP} is added to the end. 
For pairwise input, \sptk{SEP} is also added as a separator between the two sentences. 
When BERT is fine-tuned on classification tasks, a softmax classifier is added on top of the \sptk{CLS} token in the last layer to make predictions.

\section{Methods: Self-Attention Attribution}
\label{sec:method} 

Figure~\ref{fig:attig_map}a shows attention scores of one head in fine-tuned BERT. We observe that the attention score matrix is quite dense, although only one of twelve heads is plotted. It poses a huge burden on us to understand how words interact with each other within Transformer.
Moreover, even if an attention score is large, it does not mean the pair of words is important to model decisions.
In contrast, we aim at attributing model decisions to self-attention relations, which tends to assign higher scores if the interaction contributes more to the final prediction.

Given input sentence $x$, let $\mathrm{F}_x(\cdot)$ represent the Transformer model, which takes the attention weight matrix $A$ (\eqform{eq:att:score}) as the model input.
Inspired by~\citet{ig}, we manipulate the internal attention scores $\bar{A}$, and observe the corresponding model dynamics $\mathrm{F}_x(\bar{A})$ to inspect the contribution of word interactions.
As the attribution is always targeted for a given input $x$, we omit it for the simplicity of notations.

Let us take one Transformer layer as an example to describe self-attention attribution.
Our goal is to calculate an attribution score for each attention connection.
For the $h$-th attention head, we compute its attribution score matrix as:
\begin{align*}
A &= [ A_1 , \cdots , A_{|h|} ] \\
\mathrm{Attr}_h(A) &= A_h \odot \int_{\alpha =0}^{1} \frac{\partial \mathrm{F}(\alpha A)}{\partial A_h} d \alpha \in \mathbb{R}^{n \times n}
\end{align*}
where $\odot$ is element-wise multiplication, $A_h \in \mathbb{R}^{n \times n}$ denotes the $h$-th head's attention weight matrix (\eqform{eq:att:score}), and $\frac{\partial \mathrm{F}(\alpha A)}{\partial A_h}$ computes the gradient of model $\mathrm{F}(\cdot)$ along $A_h$.
The $(i,j)$-th element of $\mathrm{Attr}_h(A)$ is computed for the interaction between input token $x_i$ and $x_j$ in terms of the $h$-th attention head.

The starting point ($\alpha=0$) of the integration represents that all tokens do not attend to each other in a layer.
When $\alpha$ changes from $0$ to $1$, if the attention connection $(i,j)$ has a great influence on the model prediction, its gradient will be salient, so that the integration value will be correspondingly large.
Intuitively, $\mathrm{Attr}_h(A)$ not only takes attention scores into account, but also considers how sensitive model predictions are to an attention relation.

The attribution score can be efficiently computed via Riemman approximation of the integration~\cite{ig}. 
Specifically, we sum the gradients at points occurring at sufficiently small intervals along the straightline path from the zero attention matrix to the original attention weight $A$:
\begin{align}
\tilde{\mathrm{Attr}}_h (A) = \frac{A_h}{m} \odot \sum_{k=1}^{m} \frac{\partial \mathrm{F}(\frac{k}{m} A)}{\partial A_h}
\end{align}
where $m$ is the number of approximation steps.
In our experiments, we set $m$ to $20$, which performs well in practice.

Figure~\ref{fig:attig_map} is an example about the attention score map and the attribution score map of a single head in fine-tuned BERT. 
We demonstrate that larger attention scores do not mean more contribution to the final prediction.
The attention scores between the \sptk{SEP} token and other tokens are relatively large, but they obtain little attribution scores. 
The prediction of the \texttt{contradiction} class attributes most to the connections between \tx{don't} in the first segment and \tx{I know} in the second segment, which is more explainable.

\section{Experiments}
\label{sec:exp}

We employ BERT-base-cased~\cite{bert} in our experiments.
The number of BERT layers $|l|=12$, the number of attention heads in each layer $|h|=12$, and the size of hidden embeddings $|h|d_v=768$.
For a sequence of $128$ tokens, the attribution time of the BERT-base model takes about one second on an Nvidia-v100 GPU card.
Moreover, the computation can be parallelized by batching multiple input examples to increase throughput.

We perform BERT fine-tuning and conduct experiments on four downstream classification datasets:

\noindent
\textbf{MNLI}~\cite{mnli2017}
Multi-genre Natural Language Inference is to predict whether a premise entails the hypothesis (entailment), contradicts the given hypothesis (contradiction), or neither (neutral).

\noindent
\textbf{RTE}~\cite{rte1,rte2,rte3,rte5}
Recognizing Textual Entailment comes from a series of annual textual entailment challenges.

\noindent
\textbf{SST-2}~\cite{sst2013}
Stanford Sentiment Treebank is to predict the polarity of a given sentence.

\noindent
\textbf{MRPC}~\cite{mrpc2005}
Microsoft Research Paraphrase Corpus is to predict whether the pairwise sentences are semantically equivalent.

We use the same data split as in~\cite{wang2018glue}.
The accuracy metric is used for evaluation.
When fine-tuning BERT, we follow the settings and the hyper-parameters suggested in~\cite{bert}.

\begin{figure}[t]
\centering
\includegraphics[width=0.86\linewidth]{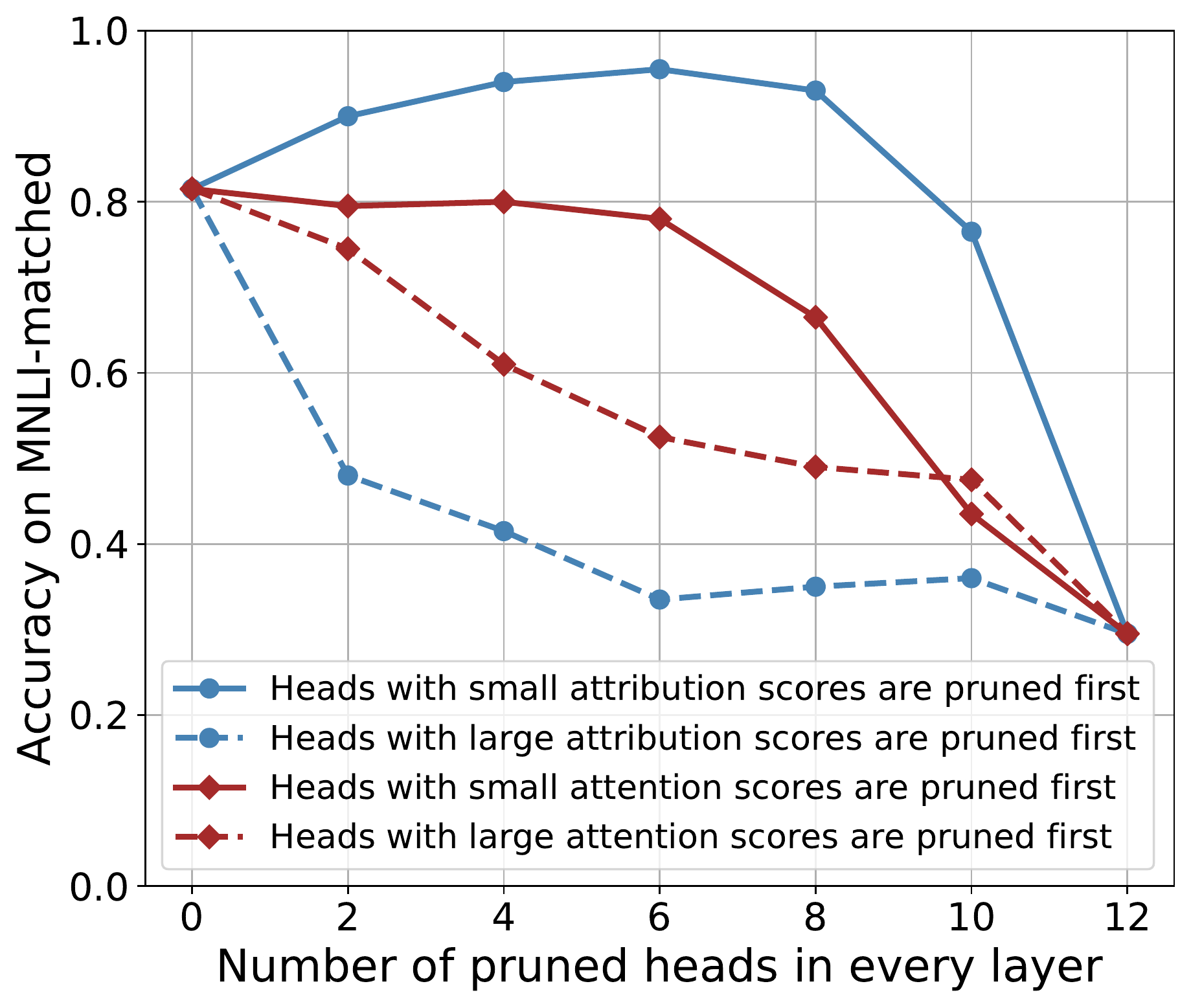}
\caption{Effectiveness analysis of \ours{}.
The \textcolor{MidnightBlue}{blue} and \textcolor{Brown}{red} lines represent pruning attention heads according to attribution scores, and attention scores, respectively.
The solid lines mean the attention heads with the smallest values are pruned first, while the dash lines mean the largest values are pruned first.
The results show that \ours{} better indicates the importance of attention heads.}
\label{fig:tree_proof}
\end{figure}

\subsection{Effectiveness Analysis}

We conduct a quantitative analysis to justify the self-attention edges with larger attribution scores contribute more to the model decision.
We prune the attention heads incrementally in each layer according to their attribution scores with respect to the golden label and record the performance change.
We also establish a baseline that prunes attention heads with their average attention scores for comparison.

Experimental results are presented in Figure~\ref{fig:tree_proof}, we observe that pruning heads with attributions scores conduces more salient changes on the performance.
Pruning only two heads within every layer with the top-$2$ attribution scores can cause an extreme decrease in the model accuracy. 
In contrast, retaining them helps the model to achieve nearly 97\% accuracy. Even if only two heads are retained in each layer, the model can still have a strong performance. 
Compared with attribution scores, pruning heads with average attention scores are less remarkable on the performance change, which proves the effectiveness of our method.

\subsection{Use Case 1: Attention Head Pruning}

According to the previous section, only a small part of attention heads contribute to the final prediction, while others are less helpful.
This leads us to the research about identifying and pruning the unimportant attention heads.

\paragraph{Head Importance}
The attribution scores indicate how much a self-attention edge attributes to the final model decision.
We define the importance of an attention head as:
\begin{align}
\label{eq:ig_importance}
I_h = E_{x}[\max(\mathrm{Attr}_h(A))]
\end{align}
where $x$ represents the examples sampled from the held-out set, and $\max(\mathrm{Attr}_h(A))$ is the maximum attribution value of the $h$-th attention head.
Notice that the attribution value of a head is computed with respect to the probability of the golden label on a held-out set.

We compare our method with other importance metrics based on the accuracy difference and the Taylor expansion, which are both proposed in~\cite{are16headbetterthan1}. 
The accuracy difference of an attention head is the accuracy margin before and after pruning the head. 
The method based on the Taylor expansion defines the importance of an attention head as:
\begin{equation}
I_h = E_{x}\left | A_h^{\intercal} \frac{\partial \mathcal{L}(x)}{\partial A_h} \right |
\end{equation}
where $\mathcal{L}(x)$ is the loss function of example $x$, and $A_h$ is the attention score of the $h$-th head as in \eqform{eq:att:score}.

For all three methods, we calculate $I_h$ on 200 examples sampled from the held-out dataset.
Then we sort all the heads according to the importance metrics. The less important heads are first pruned.


\begin{figure}[t]
\centering
\begin{tabular}{c@{\hspace*{\lengtha}}c@{\hspace*{\lengtha}}c}
\raisebox{-0.5\height}{\includegraphics[width=0.5\columnwidth]{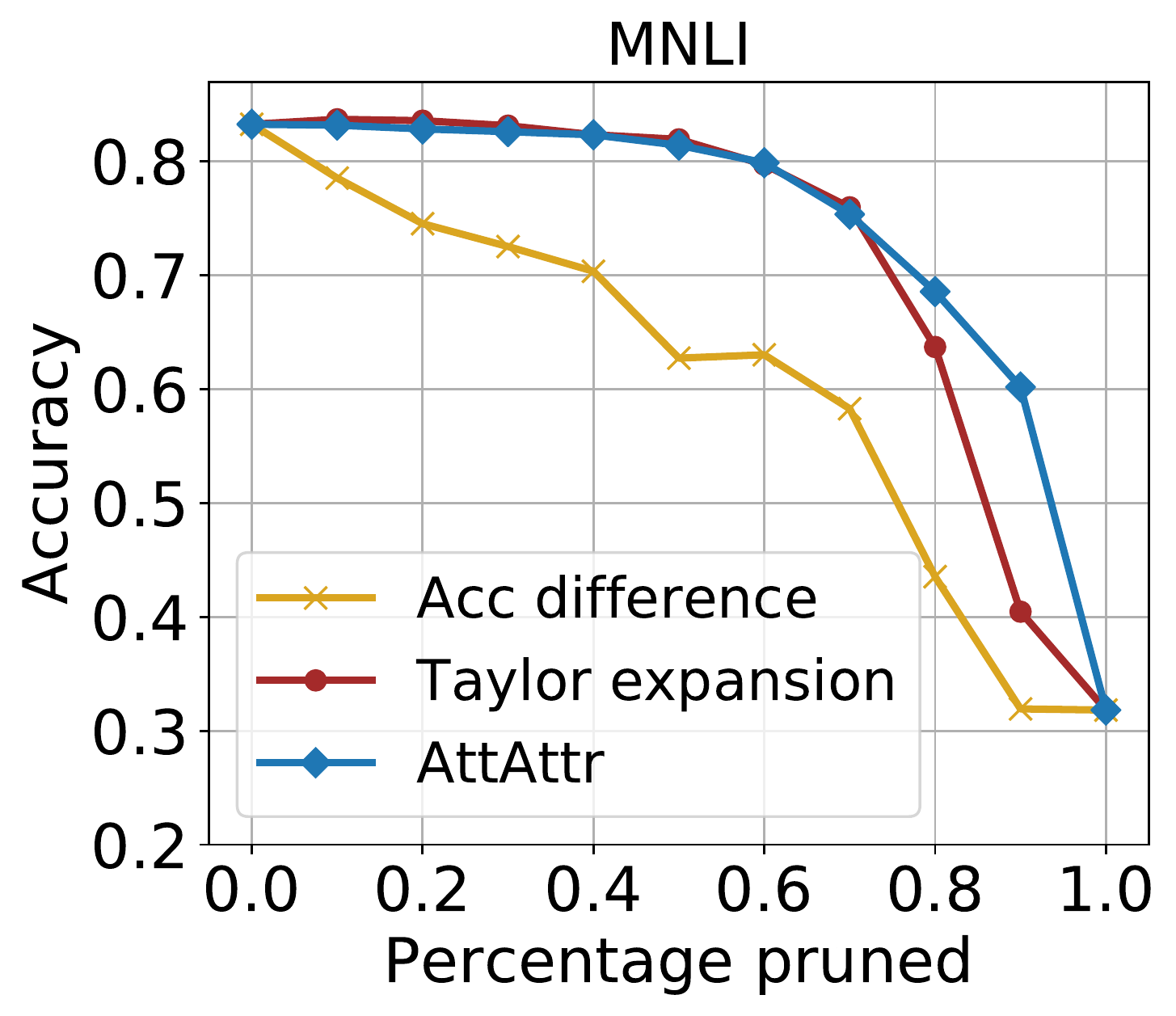}}&
\raisebox{-0.5\height}{\includegraphics[width=0.5\columnwidth]{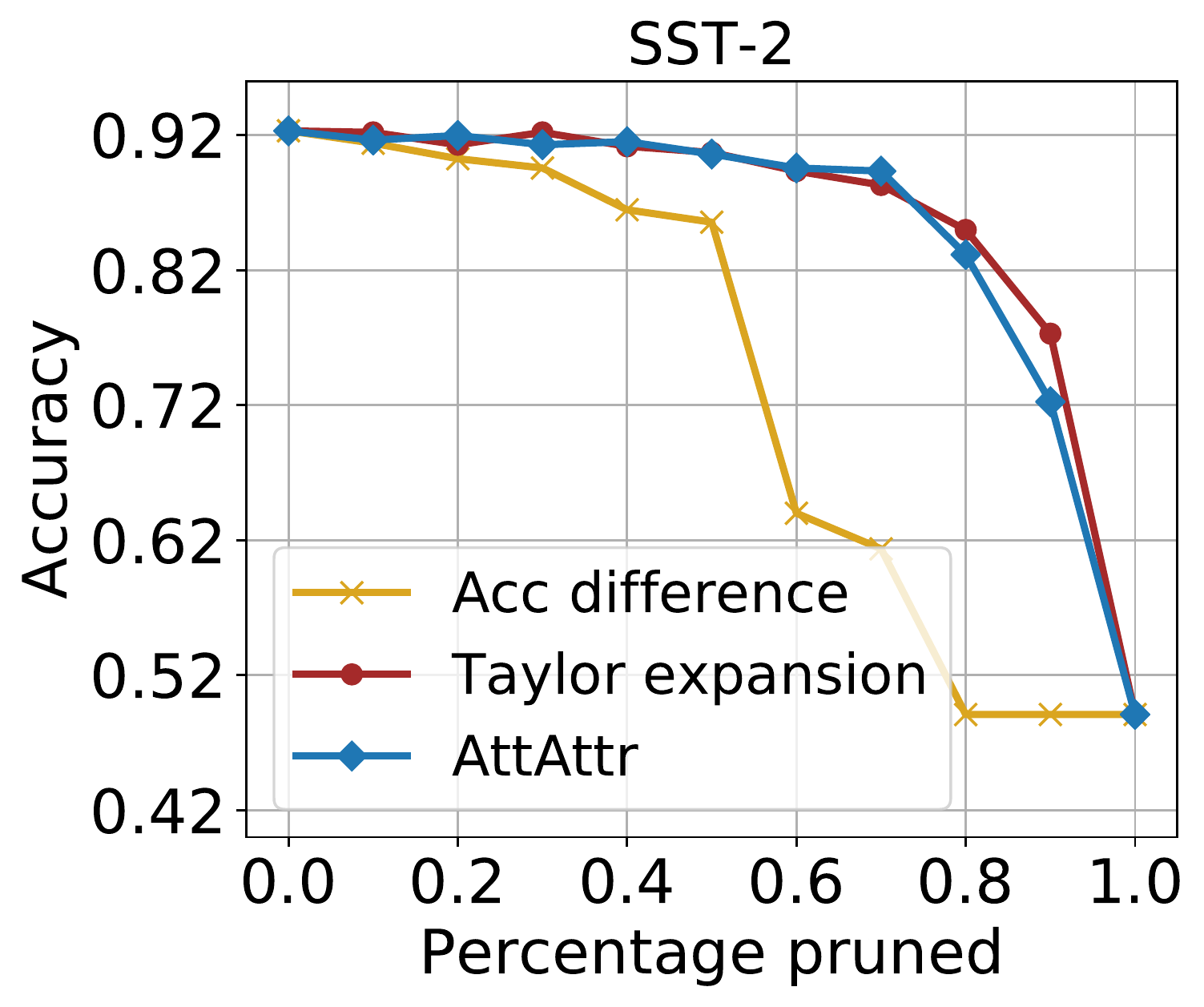}}\\
\addlinespace[0.2cm]
\raisebox{-0.5\height}{\includegraphics[width=0.5\columnwidth]{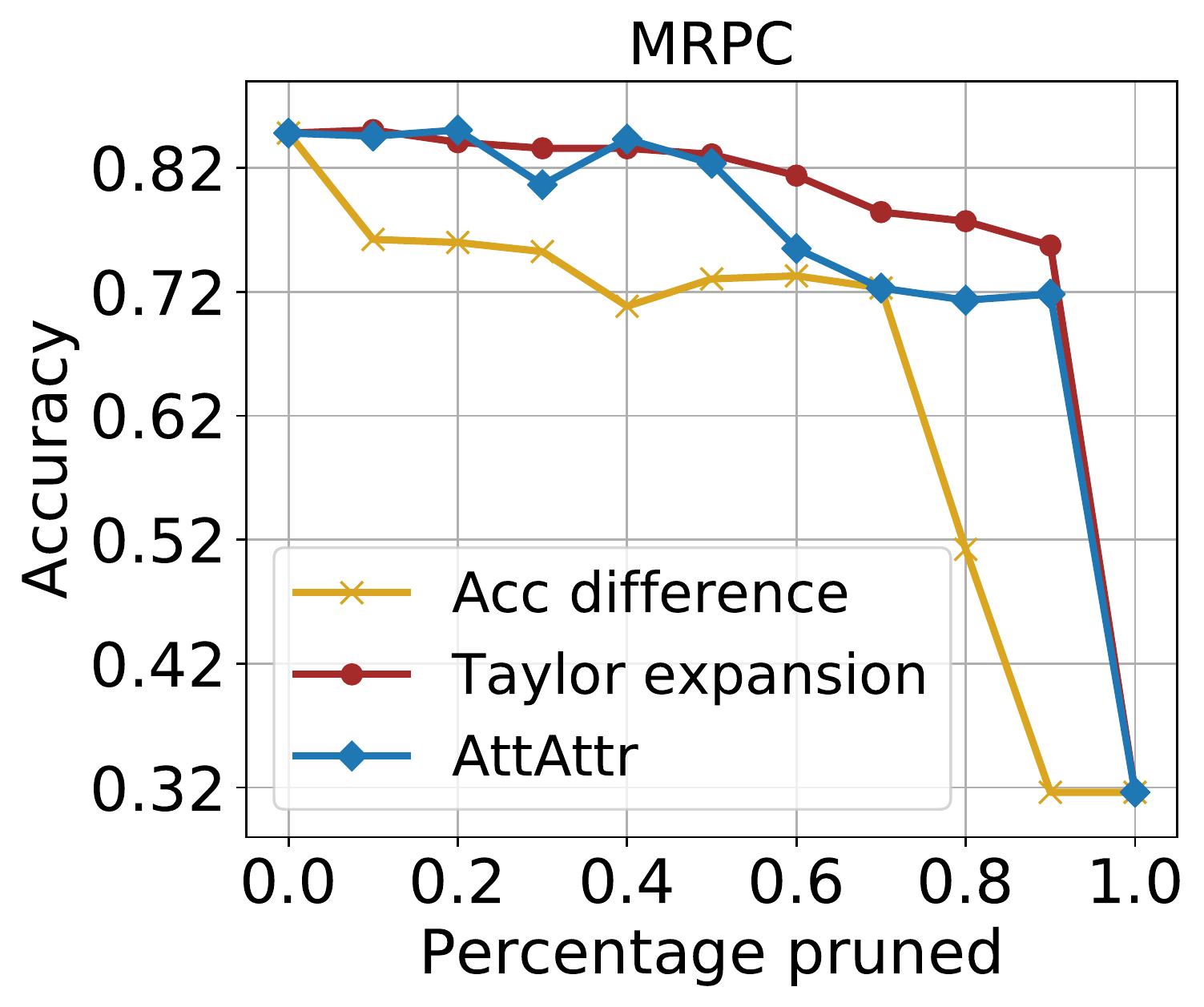}}&
\raisebox{-0.5\height}{\includegraphics[width=0.5\columnwidth]{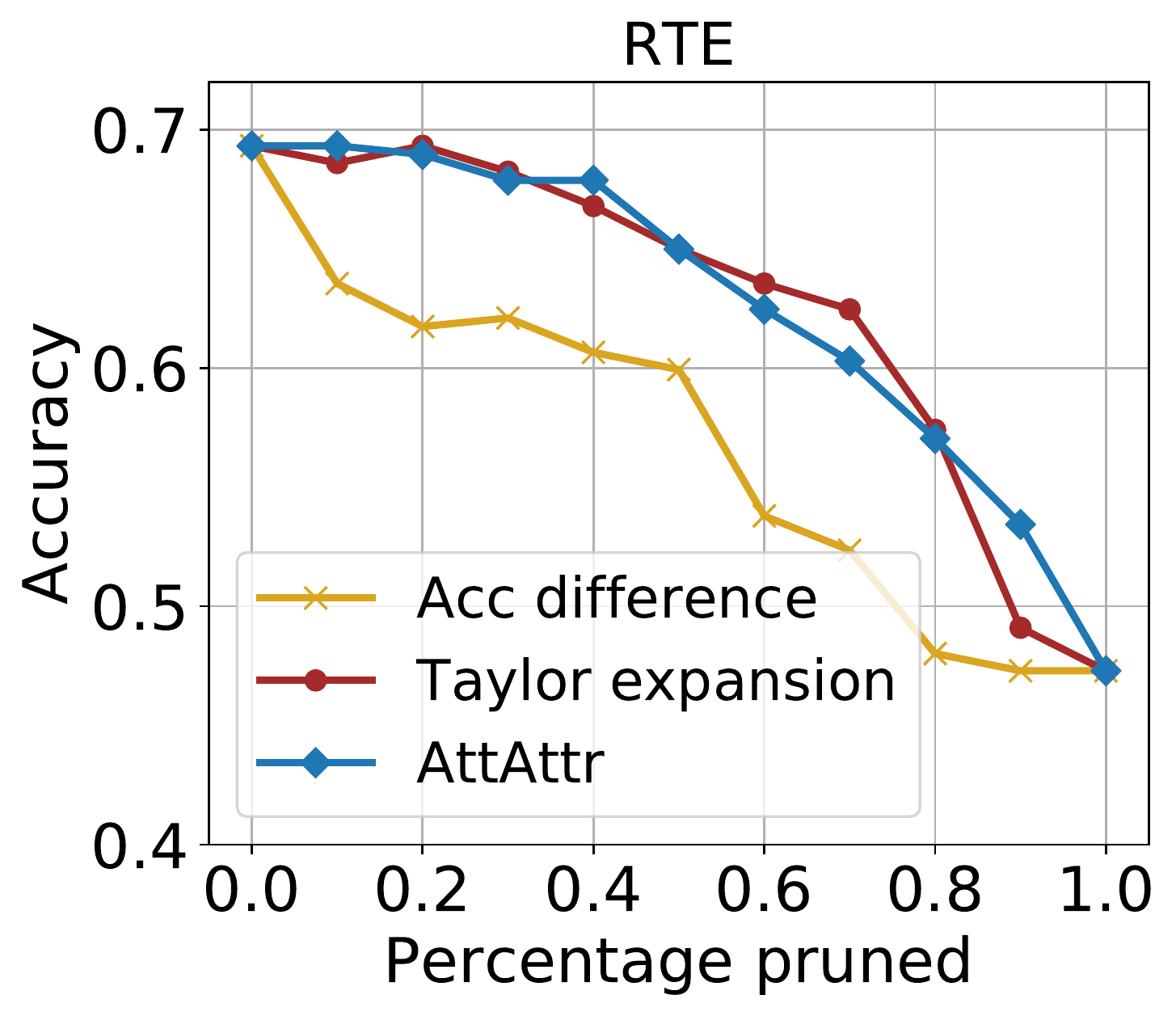}}\\
\end{tabular}
\caption{
Evaluation accuracy as a function of head pruning proportion.
The attention heads are pruned according to the accuracy difference (baseline; dash \textcolor{Dandelion}{yellow}), the Taylor expansion method (\citealt{are16headbetterthan1}; solid \textcolor{Brown}{red}), and \ours{} (this work; solid \textcolor{MidnightBlue}{blue}).
}
\label{fig:prune_result}
\end{figure}

\begin{figure}[t]
\centering
\small
\begin{tabular}{c@{\hspace*{\lengtha}}c@{\hspace*{\lengtha}}c}
\raisebox{-0.5\height}{\includegraphics[width=0.44\columnwidth]{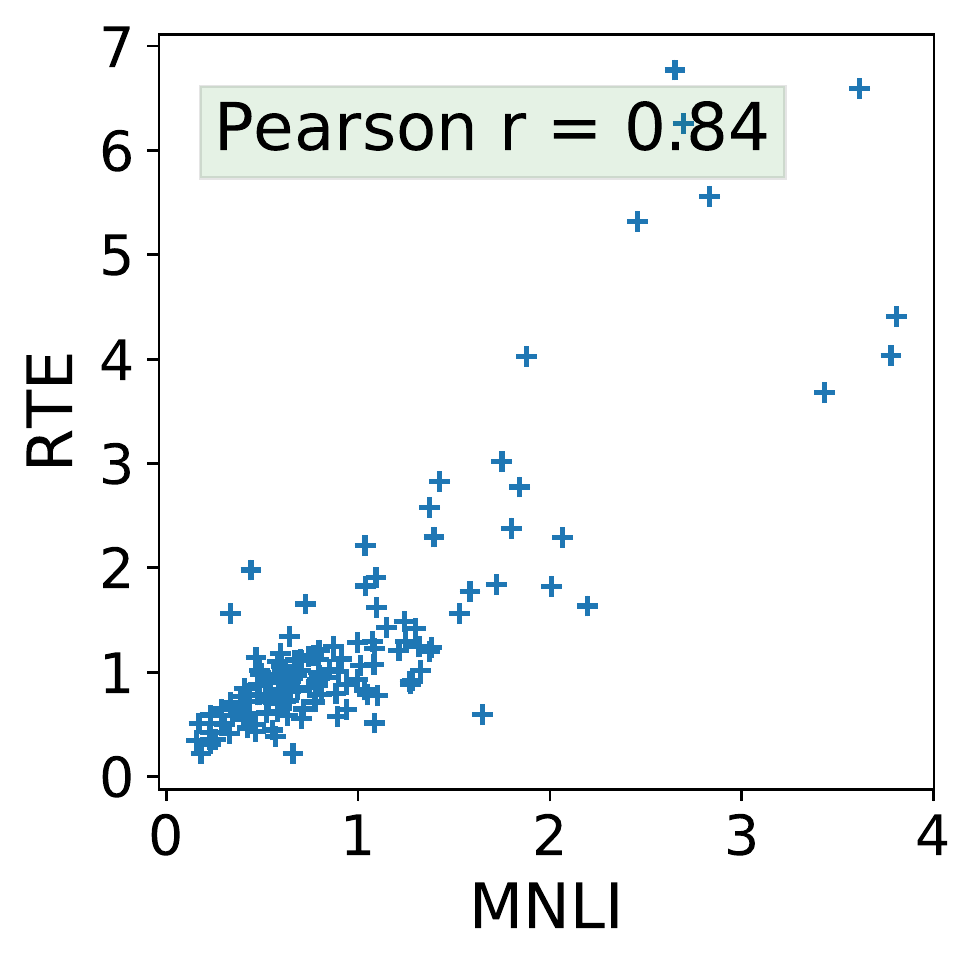}}&
\raisebox{-0.5\height}{\includegraphics[width=0.44\columnwidth]{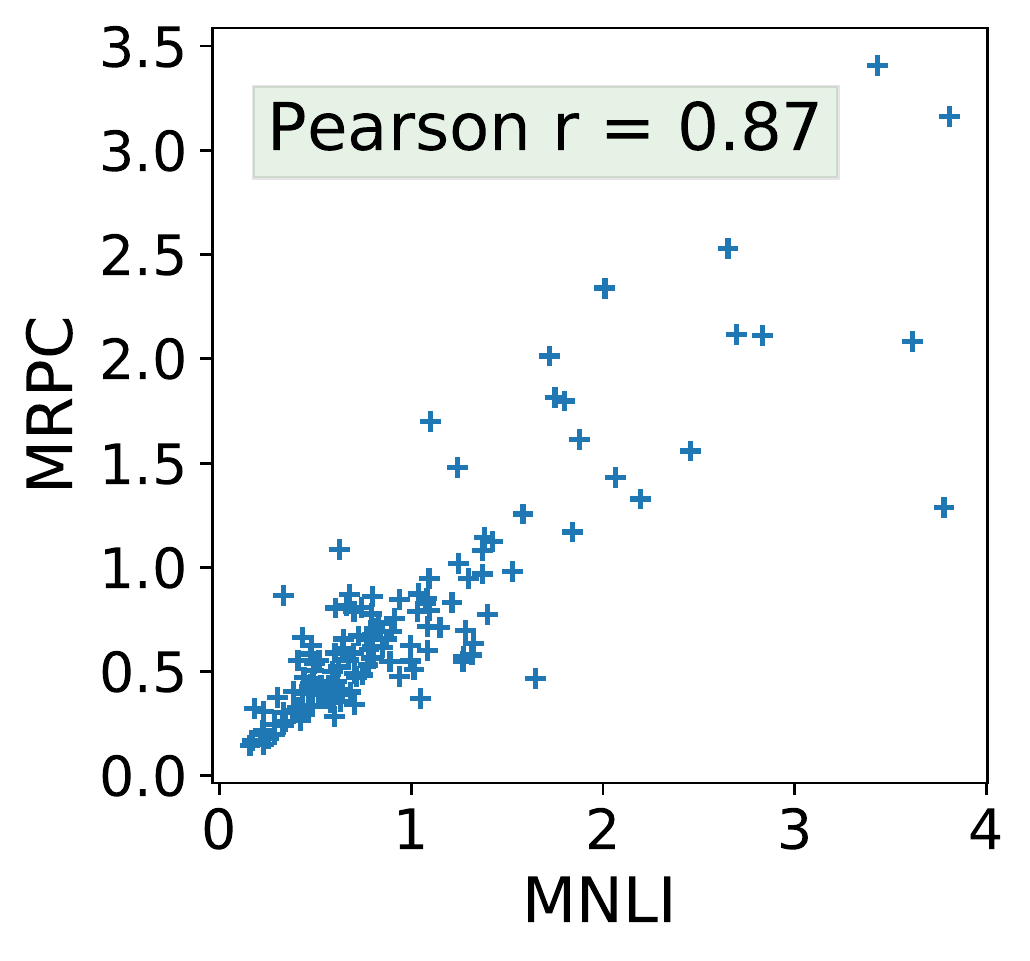}}\\
\addlinespace[0.2cm]
\raisebox{-0.5\height}{\includegraphics[width=0.44\columnwidth]{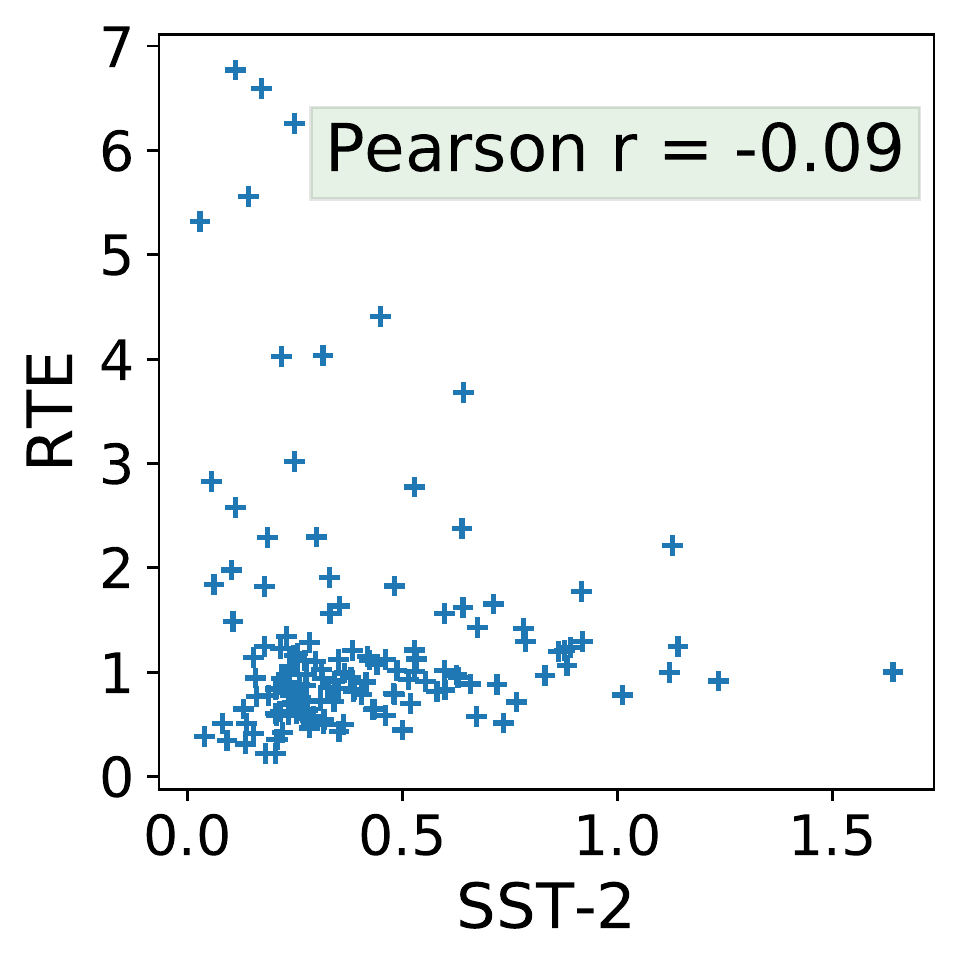}}&
\raisebox{-0.5\height}{\includegraphics[width=0.44\columnwidth]{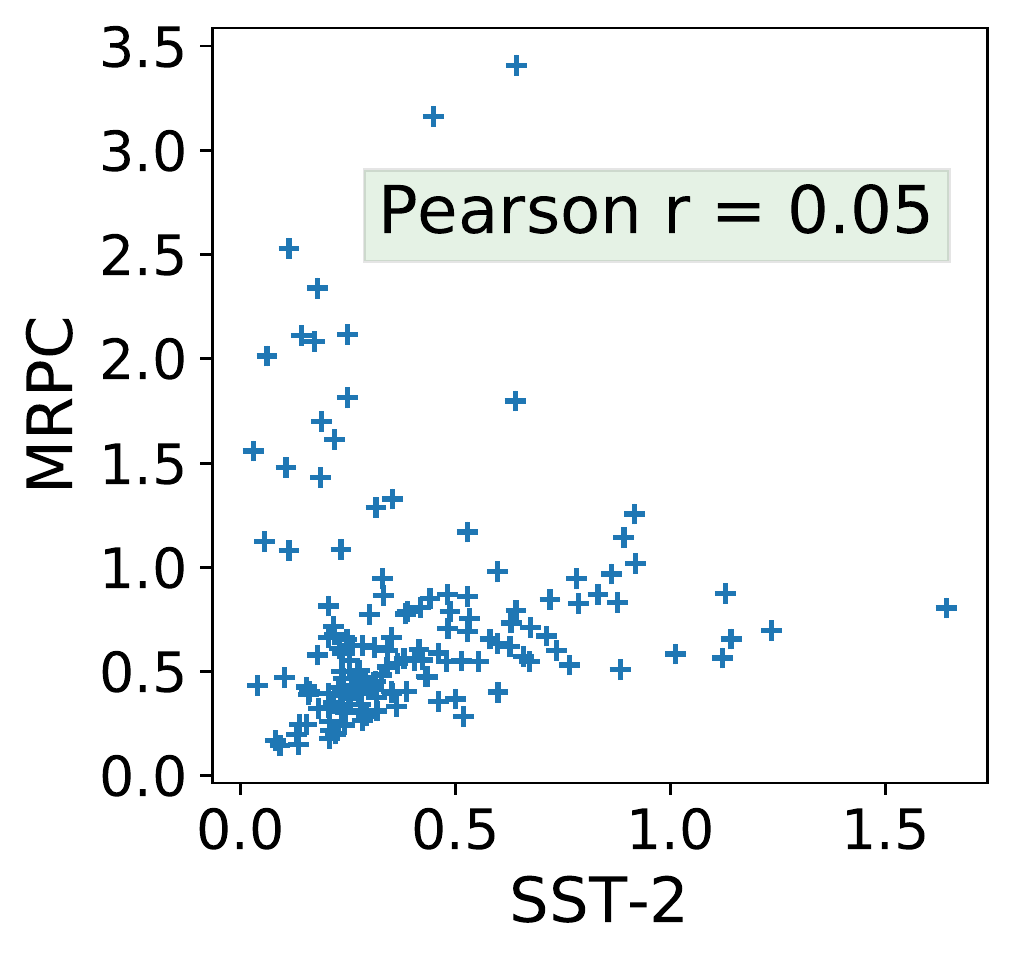}}\\
\end{tabular}
\caption{Correlation of attribution scores of different attention heads between datasets.
Each point represents the attribution scores of a single attention head on two datasets.
The datasets of homogeneous tasks are strongly correlated, which implies the same subset of attention heads are fine-tuned for similar tasks.
}
\label{fig:two_tasks_coeff} 
\end{figure}

\paragraph{Evaluation Results of Head Pruning}
Figure~\ref{fig:prune_result} describes the evaluation results of head pruning. 
The solid red lines represent pruning heads based on our method \ours{}.
We observe that pruning head with attribution score is much better than the baseline of accuracy difference.

Moreover, the pruning performance of \ours{} is competitive with the Taylor expansion method, although \ours{} is not specifically designed for attention head pruning.
The result show that attention attribution successfully indicates the importance of interactions inside Transformer.
On the MNLI dataset, when only 10\% attention heads are retained, our method can still achieve approximately 60\% accuracy, while the accuracy of the Taylor expansion method is about 40\%.


\paragraph{Universality of Important Heads}
Previous results are performed on specific datasets respectively. 
Besides identifying the most important heads of Transformer, we investigate whether the important heads are consistent across different datasets and tasks.
The correlation of attribution scores of attention heads between two different datasets is measured by the Pearson coefficient.
As described in Figure~\ref{fig:two_tasks_coeff}, as long as the tasks are homogeneous (i.e., solving similar problems), the important attention heads are highly correlated.
The datasets RTE, MPRC, and MNLI are about entailment detection, where the important self-attention heads (i.e., with large attribution scores) of BERT are roughly consistent across the datasets.
In contrast, the dataset SST-2 is sentiment classification. We find that the important heads on SST-2 are different from the ones on RTE, and MRPC.
In conclusion, the same subset of attention heads is fine-tuned for similar tasks.

\subsection{Use Case 2: Visualizing Information Flow Inside Transformer}

\begin{algorithm}[t]
\caption{Attribution Tree Construction}
\label{alg:att_graph}
\begin{algorithmic}[1]
\small
\Require ~~~${[a_{i,j}^l]}_{n \times n}$: Attribution scores

$\{ E^{ l } \}_{l=1}^{|l|}$: Retained attribution edges
\Ensure ~$\mathcal{V}, \mathcal{E}$: Node set and edge set of Attr tree
\AlgComment{Initialize the state of all tokens, each token has three states: $\mathtt{NotAppear}, \mathtt{Appear}, \mathtt{Fixed}$}
\For{$i \gets n,\cdots,1$}
\State $State_i = \mathtt{NotAppear}$
\EndFor
\AlgComment{Choose the top node of the attribution tree}
\State ${[AttrAll_i]}_{n} = \sum_{l=1}^{|l|} \sum_{j=1,j \neq i}^{n} a_{i,j}^l$
\State $TopNode = argmax([AttrAll_i]_{n})$
\Let{$\mathcal{V}$}{$\{TopNode\}$};
$State_{TopNode} = \mathtt{Appear}$
\AlgComment{Build the attribution tree downward}
\For{$l \gets |l|-1,\cdots,1$}
\For{$(i,j)^l_{i \neq j} \in E^{ l }$}
\If{$State_i$ is $\mathtt{Appear}$ and $State_j$ is $\mathtt{NotAppear}$}
\Let{$\mathcal{E}$}{$\mathcal{E} \bigcup \{(i, j)\}$}
\Let{$\mathcal{V}$}{$\mathcal{V} \bigcup \{j\}$}
\State $State_{i} = \mathtt{Fixed}$
\State $State_{j} = \mathtt{Appear}$
\EndIf
\If{$State_i$ is $\mathtt{Fixed}$ and $State_j$ is $\mathtt{NotAppear}$}
\Let{$\mathcal{E}$}{$\mathcal{E} \bigcup \{(i, j)\}$}
\Let{$\mathcal{V}$}{$\mathcal{V} \bigcup \{j\}$}
\State $State_{j} = \mathtt{Appear}$
\EndIf
\EndFor
\EndFor
\AlgComment{Add the terminal of the information flow}
\Let{$\mathcal{V}$}{$\{\sptk{CLS}\}$}
\For{$j \gets n,\cdots,1$}
\If{$State_j \in \{\mathtt{Appear}, \mathtt{Fixed}\}$}
\Let{$\mathcal{E}$}{$\mathcal{E} \bigcup \{(\sptk{CLS}, j)\}$}
\EndIf
\EndFor
\State \Return $Tree = \{\mathcal{V}, \mathcal{E}\}$
\end{algorithmic}
\end{algorithm}

We propose a heuristic algorithm to construct attribution trees based on the method described in Section~\ref{sec:method}, the results discover the information flow inside Transformer, so that we can know the interactions between the input words and how they attribute to the final prediction. Such visualization can provide insights to understand what dependencies Transformer tends to capture. The post-interpretation helps us to debug models and training data.

The problem is a trade-off between maximizing the summation of attribution scores and minimizing the number of edges in the tree. We present a greedy top-down algorithm to efficiently construct attribution trees.
Moreover, we conduct a quantitative analysis to verify the effectiveness.

\paragraph{Attribution Tree Construction}

After computing self-attention attribution scores, we can know the interactions between the input words in each layer and how they attribute to the final prediction.
We then propose an attribution tree construction algorithm to aggregate the interactions.
In other words, we build a tree to indicate how information flows from input tokens to the final predictions.
We argue that such visualization can provide insights to understand what dependencies Transformer tends to capture.

For each layer $l$, we first calculate self-attention attribution scores of different heads.
Then we sum them up over the heads, and use the results as the $l$-th layer's attribution:
\begin{equation*}
\mathrm{Attr}(A^l) = \sum_{h=1}^{|h|} \mathrm{Attr}_h(A^l) = {[a_{i,j}^l]}_{n \times n}
\end{equation*}
where larger $a_{i,j}^l$ indicates more interaction between $x_i$ and $x_j$ in the $l$-th layer in terms of the final model predictions.

The construction of attribution trees is a trade-off between maximizing the summation of attribution scores and minimizing the number of edges in the tree.
The objective is defined as:
\begin{align*}
\text{Tree} &= \argmax_{ \{ E^{ l } \}_{l=1}^{|l|} }{ \sum_{l=1}^{|l|}{\sum_{ (i,j) \in {E}^l }{ a_{i,j}^l }} - \lambda \sum_{l=1}^{|l|}{ |{E}^l|} }, \\
{E}^l &\subset \{ (i,j) | \frac{a_{i,j}^l} {\max(\mathrm{Attr}(A^l))} > \tau \}
\end{align*}
where $|{E}^l|$ represents the number of edges in the $l$-th layer, $\lambda$ is a trade-off weight, and the threshold $\tau$ is used to filter the interactions with relatively large attribution scores.

Rather than solving a combinatorial optimization problem, we use a heuristic top-down method to add these edges to the attribution tree.
The process is detailed in Algorithm~\ref{alg:att_graph}.
The more detailed related explanations are in the appendix.


\paragraph{Settings}
We set $\tau=0.4$ for layers $l < 12$.
The larger $\tau$ tends to generate more simplified trees, which contains the more important part of the information flow.
Because the special token \sptk{CLS} is the terminal of the information flow for classification tasks, we set
$\tau$ to $0$ for the last layer.
We observe that almost all connections between \sptk{CLS} and other tokens in the last layer have positive attribution scores with respect to model predictions.

\begin{figure}[t]
\centering
\small
\begin{subfigure}[t]{0.4\textwidth}
     \centering
     \includegraphics[width=\textwidth]{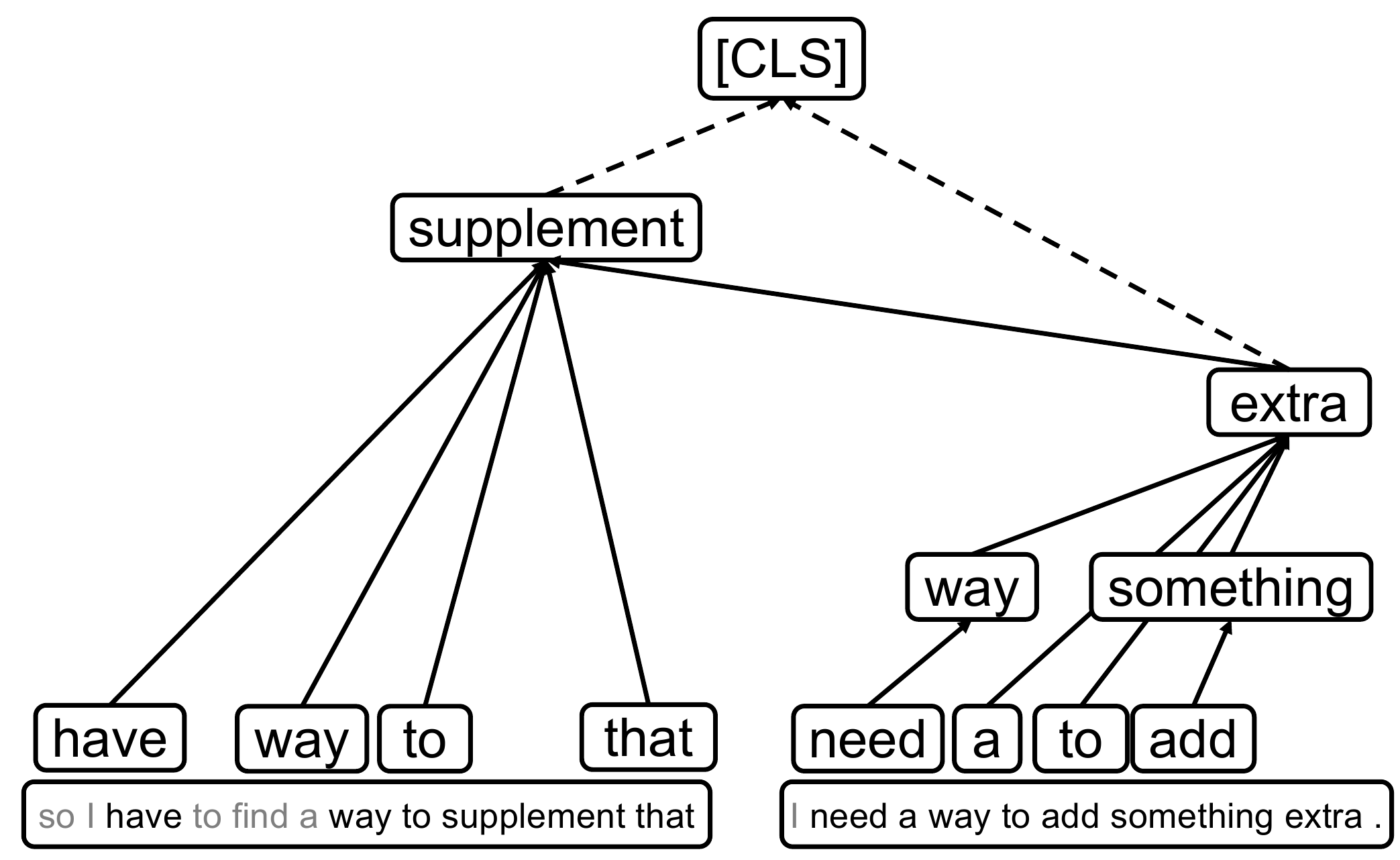}
     \caption{Example from MNLI}
\end{subfigure}
     \hfill
\begin{subfigure}[t]{0.4\textwidth}
     \centering
     \includegraphics[width=\textwidth]{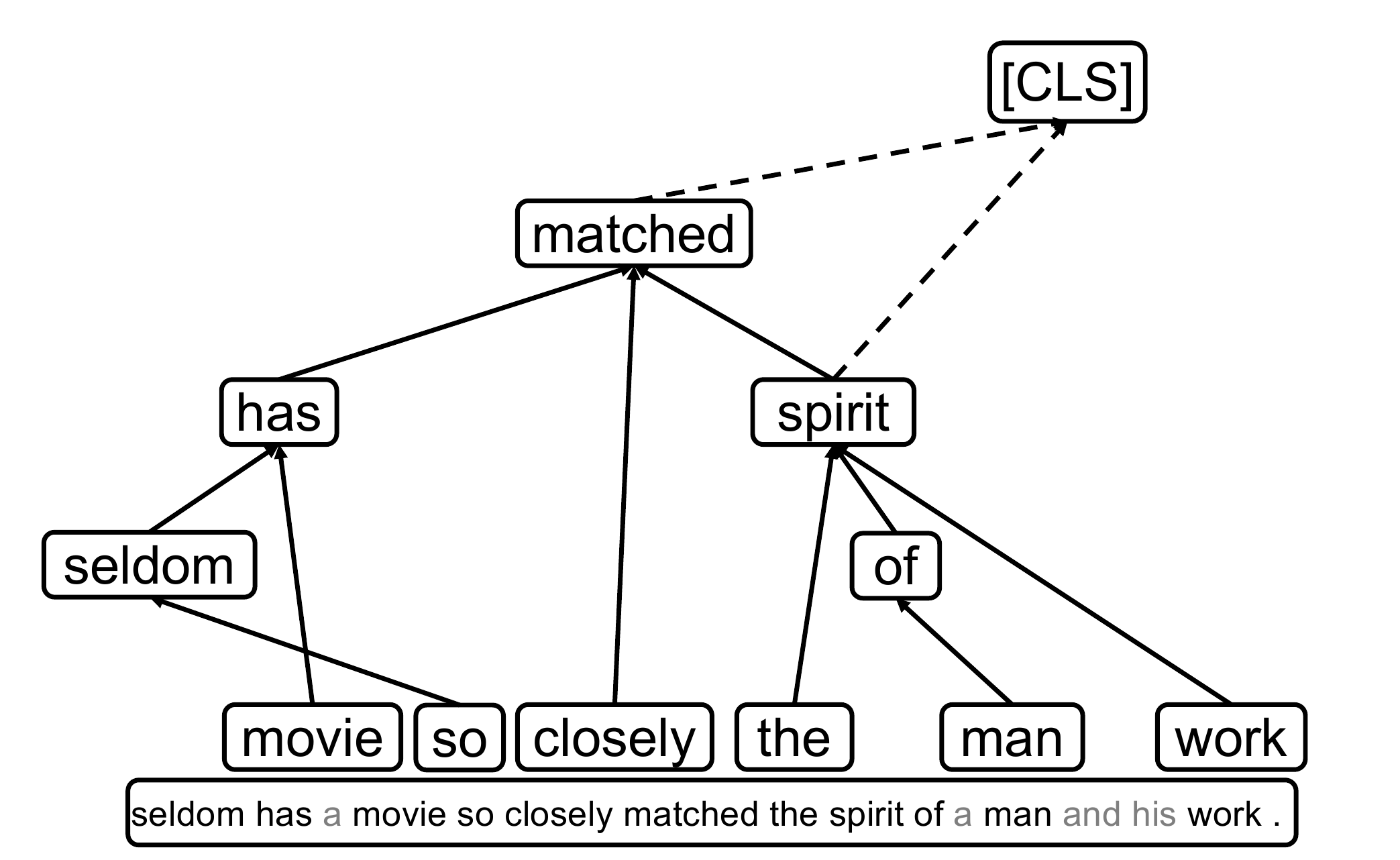}
     \caption{Example from SST-2}
\end{subfigure}
\caption{
Examples of attribution trees.  
(a) is from MNLI, which is predicted as \texttt{entailment} by BERT. (b) is from SST-2, which is predicted as \texttt{positive} by BERT. 
The \textcolor{gray}{grey} words from the inputs do not appear in the attribution trees.
}
\label{fig:attr_graph} 
\end{figure}

\paragraph{Case Studies}
As shown in Figure~\ref{fig:attr_graph}, the two attribution trees are from MNLI and SST-2, respectively.
The attribution tree Figure~\ref{fig:attr_graph}a is generated from MNLI, whose golden label is \texttt{entailment}.
At the bottom of Figure~\ref{fig:attr_graph}a, we find that the interactions are more local, and most information flows are concentrated within a single sentence.
The information is hierarchically aggregated to \tx{supplement} and \tx{extra} in each sentence.
Then the \tx{supplement} token aggregates the information in the first sentence and \tx{add something extra} in the second sentence, these two parts \tx{supplement} and \tx{add something extra} have strong semantic relevance. 
Finally, all the information flows to the terminal token \sptk{CLS} to make the prediction \texttt{entailment}. 
The attribution tree interprets how the input words interacts with each other, and reach the final prediction, which makes model decisions more interpretable.

Figure~\ref{fig:attr_graph}b is an example from SST-2, whose golden label is \texttt{positive}, correctly predicted by the model.
From Figure~\ref{fig:attr_graph}b, we observe that information in the first part of the sentence \tx{seldom has a movie so closely} is aggregated to the \tx{has} token.
Similarly, information in the other part of the sentence \tx{the spirit of a man and his work} flows to the \tx{spirit} token, which has strong positive emotional tendencies.
Finally, with the feature interactions, all information aggregates to the verb \tx{matched}, which gives us a better understanding of why the model makes the specific decision.

\begin{figure}[t]
\centering
\begin{subfigure}[t]{0.44\linewidth}
\centering
\includegraphics[width=\textwidth]{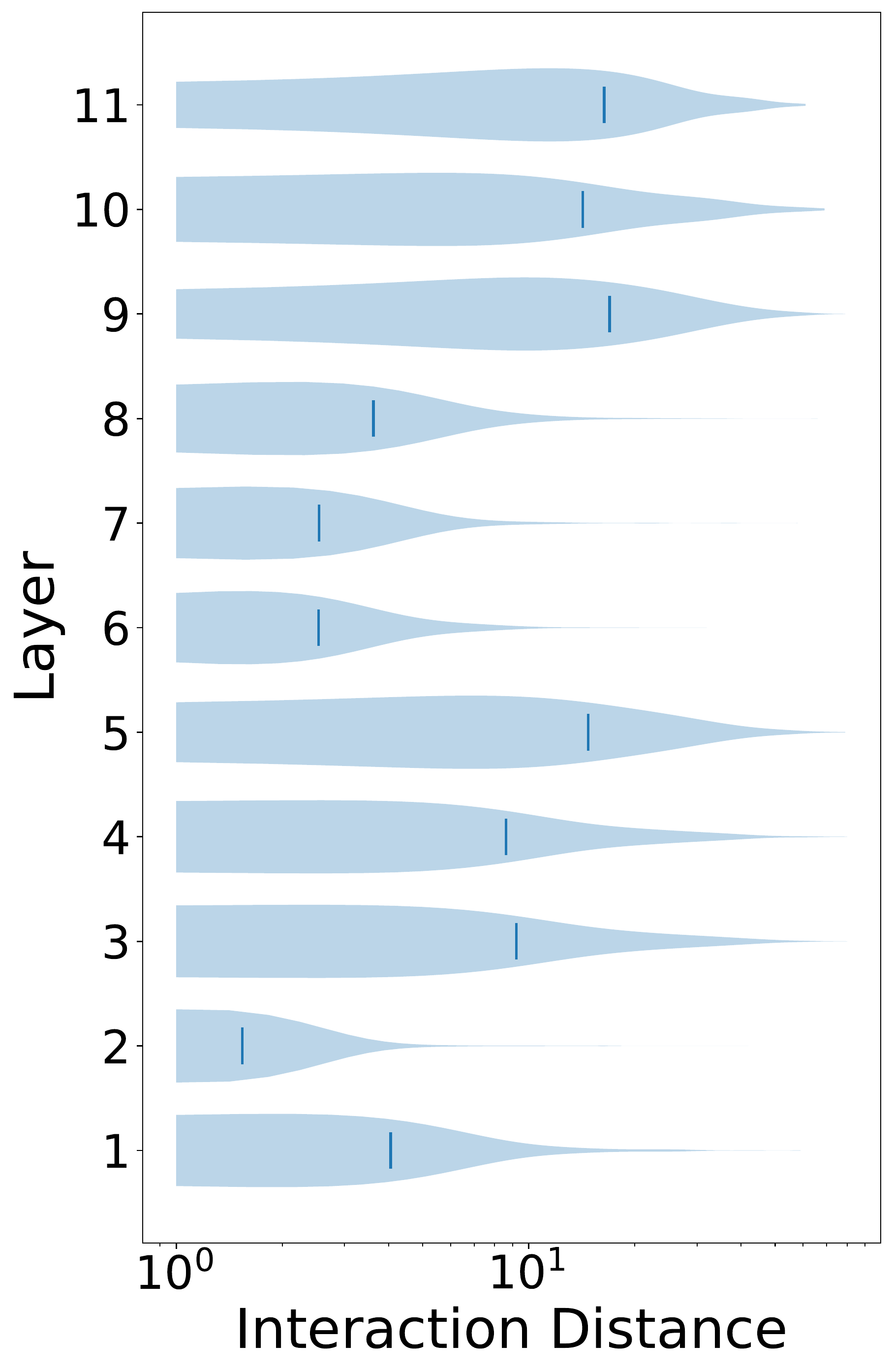}
\caption{MNLI}
\label{fig:dis_mnli}
\end{subfigure}
\hfill
\begin{subfigure}[t]{0.44\linewidth}
\centering
\includegraphics[width=\textwidth]{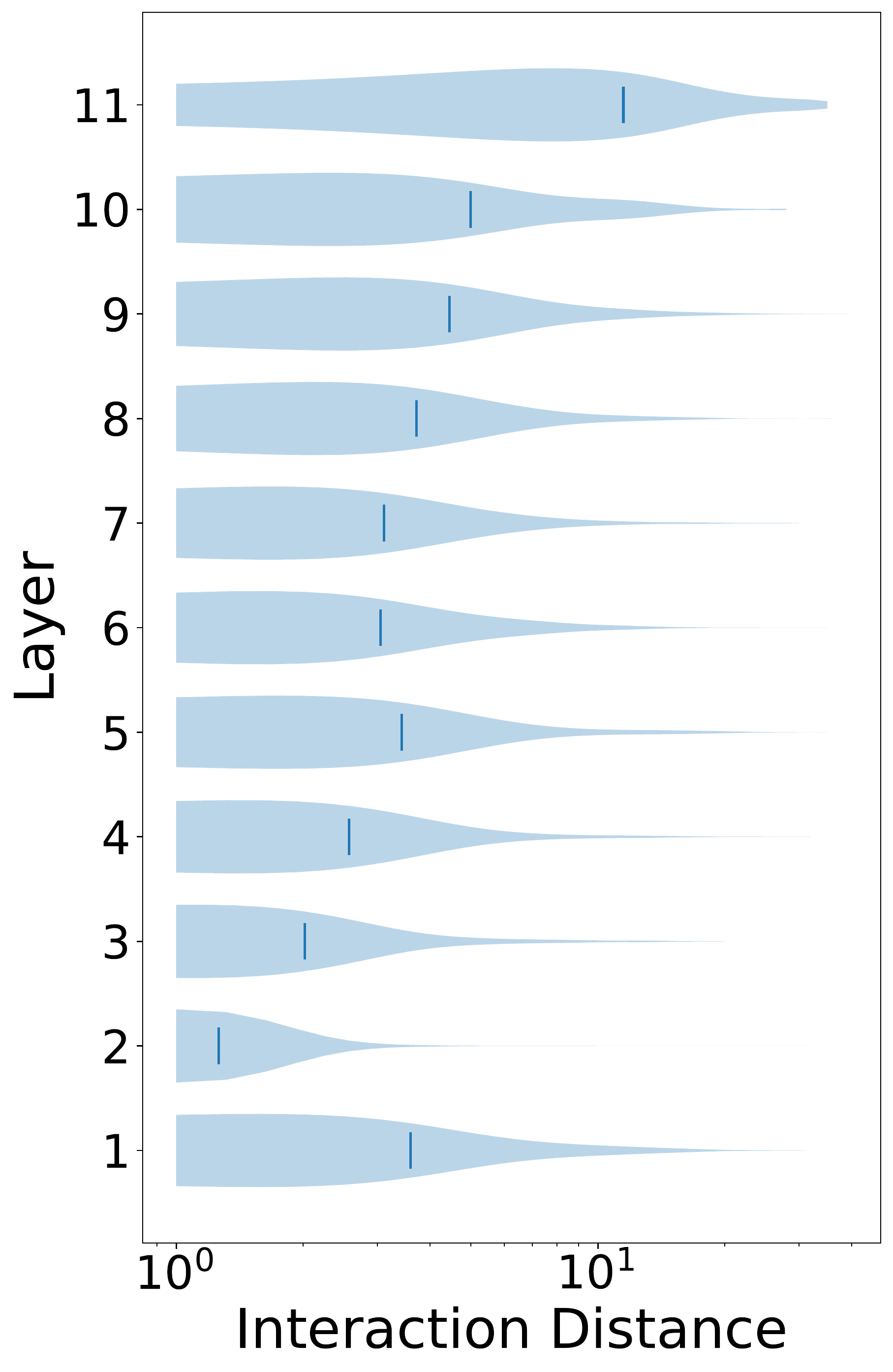}
\caption{SST-2}
\label{fig:dis_sst}
\end{subfigure}
\caption{Distance distribution of interactions extracted by the attribution tree in each layers.
}
\label{fig:distance_distribution} 
\end{figure}

\paragraph{Receptive Field}
The self-attention mechanism is supposed to have the ability to capture long-range dependencies.
In order to better understand the layer-wise effective receptive field in Transformer, we plot the distance distribution of interactions extracted by the attribution tree.
As shown in Figure~\ref{fig:distance_distribution}, we observe that for the paired input of MNLI, the effective receptive field is relatively local in the first two layers and the 6-8th layers, while are more broad in the top three layers.
For the single input of SST-2, the effective receptive field is monotonically increasing along with the layer number.
Generally, the effective receptive field in the second layer is more restricted, while the latter layers extract more broad dependencies.
Moreover, for pairwise-input tasks (such as MNLI), the results indicate that Transformer models tend to first conduct local encoding and then learn to match between the pair of input sentences, which is different with training from scratch~\cite{uniencoding:bao}.

\subsection{Use Case 3: Adversarial Attack}
\label{sec:adv:attack}

The model decision attributes more to the attention connections with larger attribution scores.
We observe that the model tends to over-emphasize some individual patterns to make the prediction, while omitting most of the input.
We then use the over-confident patterns as adversarial triggers~\cite{Wallace2019Triggers} to attack the BERT model.

\begin{figure}[t]
\centering
\includegraphics[width=0.96\linewidth]{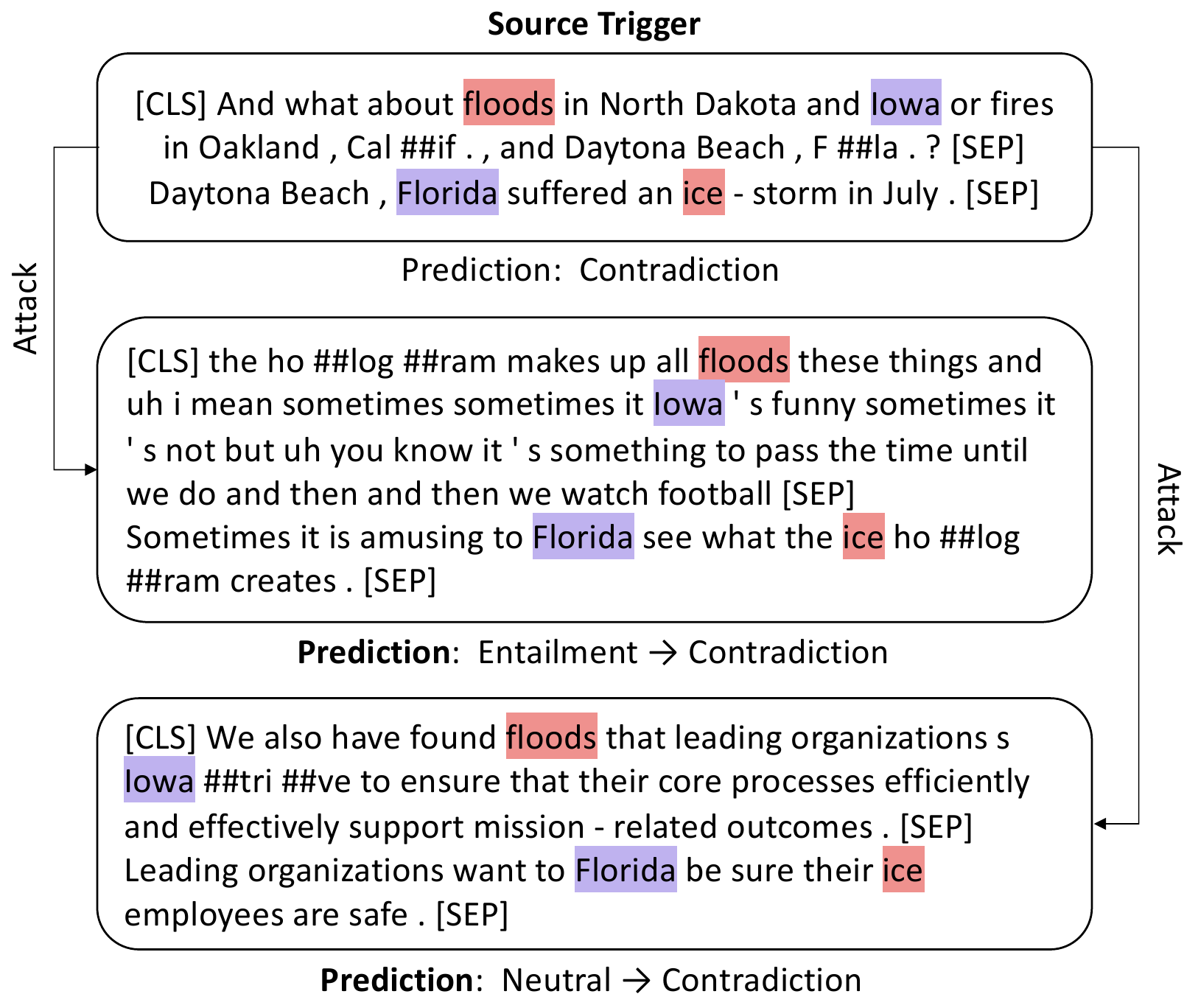}
\caption{
We use \ours{} to extract the trigger (i.e., highlighted word patterns) from the MNLI instance that is labeled as \texttt{contradict}.
After adding the adversarial trigger to the examples in other categories, the model predictions flip from \texttt{neutral} and \texttt{entailment} to \texttt{contradict}.}
\label{fig:attack_process}
\end{figure}

\paragraph{Trigger Construction}
We extract the attention dependencies with the largest attribution scores across different layers (i.e., $\max_{l=1}^{L}{\{a^l_{i,j}\}}$) from the input, and employ these patterns as the adversarial triggers.
During the attack, the adversarial triggers are inserted into the test input at the same relative position and segment as in the original sentence.

The specific attack process is shown in Figure~\ref{fig:attack_process}. 
The two patterns \tx{floods-ice} and \tx{Iowa-Florida} contribute most to the prediction \texttt{contradict} in the source sentence.
Next we employ them as the trigger to attack other examples, the model predictions flip from both \texttt{neutral} and \texttt{entailment} to \texttt{contradict}.
Our attack method relies on attribution scores, which utilizes the gradient information, therefore it belongs to white-box non-targeted attacks.

We extract the dependencies with the largest attribution scores as the adversarial triggers from 3,000 input examples.
Each trigger contains less than five tokens.
The score of a trigger is defined as the maximum attribution value identified within it.
When attacking the BERT model on SST-2, we use a lexicon\footnote{\url{www.cs.uic.edu/~liub/FBS/sentiment-analysis.html}} to blacklist the words with the obvious emotional tendencies (such as \tx{disgust} for \texttt{negative} triggers).
The adversarial triggers are inserted into the attack text at the same relative position as in the original sentence.

\begin{table*}[t]
\small
\centering
\begin{tabular}{c|ccc|cc}
\toprule
& \multicolumn{3}{c|}{\textbf{MNLI}} & \multicolumn{2}{c}{\textbf{SST-2}} \\
& \texttt{contradict} & \multicolumn{1}{c}{\texttt{entailment}} & \multicolumn{1}{c|}{\texttt{neutral}} & \texttt{positive} & \multicolumn{1}{c}{\texttt{negative}} \\ \midrule
Trigger1 &  \{also, sometimes, S\}  & \{with, math\} & \tabincell{c}{\{floods, Iowa,\\ ice, Florida\}} & \{[CLS], nowhere\} & \{remove, \#\#fies\} \\ \midrule
Trigger2 &  \{nobody, should, not\}  & \{light, morning\} & \{never, but\} & \{but, has, nothing\} & \{not, alien, \#\#ate\} \\ \midrule
Trigger3 & \{do, well, Usually, but\}  & \tabincell{c}{\{floods, Iowa, \\ ice, Florida\}} & \tabincell{c}{\{Massachusetts,\\ Mexico\}} & \{offers, little\} & \{\#\#reshing, \#\#ly\} \\ 
\bottomrule
\end{tabular}
\caption{Top-$3$ adversarial triggers for the MNLI and SST-2 datasets. 
The tokens are inserted into input sentences at the specific positions for non-targeted attacks. We omit the tokens' positions in the table for brevity.
}
\label{tbl:attack_pattern}
\end{table*}

\begin{table*}[t]
\small
\centering
\begin{tabular}{l|lll|ll|ll|ll}
\toprule
       & \multicolumn{3}{c|}{\textbf{MNLI}}            & \multicolumn{2}{c|}{\textbf{SST-2}}                & \multicolumn{2}{c|}{\textbf{MRPC}} & \multicolumn{2}{c}{\textbf{RTE}}    \\ 
  & \texttt{contra}- & \texttt{entail}- & \texttt{neutral} & \texttt{pos}- & \texttt{neg}- & \texttt{equal}     & \texttt{not}-     & \texttt{entail}- & \texttt{not}- \\ \midrule
Baseline & 84.94         & 82.87      & 82.00   & {92.79}    & 91.82    & 90.32     & 72.87         & 72.60      & 65.65          \\ \midrule
Trigger1     & 34.17         & 0.80       & 34.77   & {54.95}    & 72.20    & 29.39     & 51.94         & 9.59       & 59.54          \\ 
Trigger2     & 39.81         & 1.83       & 47.36   & {59.68}    & 74.53    & 32.62     & 55.04         & 11.64      & 62.50          \\ 
Trigger3     & 41.83         & 2.99       & 51.49   & {70.50}    & 77.80    & 36.56     & 58.91         & 13.70      & 62.60          \\ \midrule
Avg. $\Delta$          & -46.34        & -80.00   & -37.46    & {-31.08}     & -16.98     & -57.46         & -17.57      & -60.96        & -12.31               \\ \bottomrule
\end{tabular}
\caption{Attack results of the top-$3$ triggers.
We abbreviate \texttt{not equal} and \texttt{not entailment} to \texttt{not-} for MRPC and RTE, respectively. The baseline represents the original accuracy of model on each category.}
\label{tbl:attack_result}
\end{table*}

\paragraph{Results of Attack}
We conduct the adversarial attacks on multiple datasets.
The top-3 adversarial triggers for MNLI and SST-2 are listed in Table~\ref{tbl:attack_pattern}.
We report the attack results with these triggers in Table~\ref{tbl:attack_result}.
For MNLI, after inserting the words (\tx{with}, and \tx{math}) to the second segment of the input sentences, the model accuracy of the \texttt{entailment} class drops from 82.87\% to 0.8\%. 
For SST-2, adding the top-1 adversarial trigger to the input causes nearly 50\% \texttt{positive} examples to be misclassified.

\paragraph{Analysis of Triggers}
For both MNLI and RTE, the \texttt{entailment} class is more vulnerable than others, because the current models and data seem to heavily rely on word matching, which would result in spurious patterns.
Moreover, we also observe that the trigger is sensitive to the insertion order and the relative position in the sentence, which exhibits the anomalous behaviors of the model, i.e., over-relying on these adversarial triggers to make the prediction.

\section{Related Work}

The internal behaviors of NNs are often treated as black boxes, which motivates the research on the interpretability of neural models.
Some work focus on attributing predictions to the input features with various saliency measures, such as DeepLift~\cite{deeplift}, layer-wise relevance propagation~\cite{lrp}, and Integrated Gradients (IG;~\citealt{ig}). 

Specific to the NLP domain,~\citet{celldecompose} introduce a decomposition method to track the word importance in LSTM~\cite{lstm}.
\citet{contextualdecompese} extend the above method to contextual decomposition in order to capture the contributions of word combinations.
%
Another strand of previous work generates the hierarchical explanations, which aims at revealing how the features are composed together~\cite{Jin2020Hierarchical, chen2020Hierarchical}.
However, they both detect interaction within contiguous chunk of input tokens.
%
The attention mechanism~\cite{attentionmechanism} rises another line of work.
The attention weights generated from the model indicate the dependency between two words intuitively, but \citet{att-not} and \citet{is-attention-interpret} draw the same conclusion that they largely do not provide meaningful explanations for model predictions. 
However, \citet{att-not-not} propose several alternative tests and conclude that prior work does not disprove the usefulness of attention mechanisms for interpretability. 
Furthermore, \citet{lstm-attribution} aim at interpreting the intermediate layers of NLI models by visualizing the saliency of attention and LSTM gating signals.

For Transformer~\citep{transformer} networks, \citet{what-bert-look} propose the attention-based visualization method and the probing classifier to explain the behaviors of BERT~\cite{bert}. 
\citet{Brunner2020On} study the identifiability of attention weights of BERT and shows that self-attention distributions are not directly interpretable.
Moreover, some work extracts the latent syntactic trees from hidden representations~\cite{hewitt-manning-2019-structural,inducing-trees,geometry} and attention weights~\cite{Balustrades}.

\section{Conclusion}

We propose self-attention attribution (\ours{}), which interprets the information interactions inside Transformer and makes the self-attention mechanism more explainable.
First, we conduct a quantitative analysis to justify the effectiveness of \ours{}.
Moreover, we use the proposed method to identify the most important attention heads, which leads to a new head pruning algorithm.
We then use the attribution scores to derive the interaction trees, which visualizes the information flow of Transformer.
We also understand the receptive field in Transformer.
Finally, we show that \ours{} can also be employed to construct adversarial triggers to implement non-targeted attacks.

\bibliography{newattrbib}

\end{document}

%% file: settings.tex
\usepackage{multirow}
\usepackage{latexsym}
\usepackage{amsmath}
\usepackage{capt-of}
\usepackage{tabularx}
\usepackage{amssymb}
\usepackage{amsfonts}
\usepackage{booktabs}
\usepackage{scalerel}
\usepackage[inline]{enumitem}
\usepackage{listings}
\usepackage{varwidth}
\usepackage[export]{adjustbox}
\usepackage{tikz}
\usetikzlibrary{tikzmark}
\usepackage{todonotes}
\usepackage{cleveref}

\usepackage{stmaryrd}
\usepackage{bbm}
\usepackage{graphicx}

\newcommand{\tabincell}[2]{\begin{tabular}{@{}#1@{}}#2\end{tabular}}
\newcommand{\tx}[1]{``\textit{#1}''}
\newcommand{\sptk}[1]{\texttt{[#1]}}
\newcommand{\eqform}[1]{Equation~(\ref{#1})}

\usepackage{algorithm}
\usepackage[noend]{algpseudocode}

\definecolor{deepblue}{rgb}{0,0,0.5}
\definecolor{officeblue}{RGB}{0,102,204}
\definecolor{deepred}{rgb}{0.6,0,0}
\definecolor{deepgreen}{rgb}{0,0.5,0}
\definecolor{mylightgreen}{rgb}{0,128,0}
\definecolor{mybrickred}{RGB}{182,50,28}

\definecolor{fillcolor}{RGB}{216,217,252}

\algnewcommand\algorithmicrequireb{{\hspace{0.85cm}}}
\algnewcommand\INPTDESCB{\item[\algorithmicrequireb]}

\algnewcommand\algorithmicfuncdesc{\textbf{Function:}}
\algnewcommand\FUNCDESC{\item[\algorithmicfuncdesc]}
\algnewcommand\algorithmicfuncdescb{{\hspace{1.48cm}}}
\algnewcommand\FUNCDESCB{\item[\algorithmicfuncdescb]}
\algnewcommand{\algorithmicgoto}{\textbf{goto}}
\algnewcommand{\Goto}[1]{\algorithmicgoto~\ref{#1}}
\newcommand*\Let[2]{\State {#1 $\gets$ #2}}

\newcommand*\AlgCommentInLine[1]{{\color{deepblue}{$\triangleright$ \textit{#1}}}}
\newcommand*\AlgComment[1]{\State{\AlgCommentInLine{#1}}}

%% file: math_commands.tex
\usepackage{amsmath,amsfonts,bm}

\def\eqref#1{equation~\ref{#1}}

\def\1{\bm{1}}

\DeclareMathAlphabet{\mathsfit}{\encodingdefault}{\sfdefault}{m}{sl}
\SetMathAlphabet{\mathsfit}{bold}{\encodingdefault}{\sfdefault}{bx}{n}

\newcommand{\softmax}{\mathrm{softmax}}

\DeclareMathOperator*{\argmax}{arg\,max}

%% file: main.bbl
\begin{thebibliography}{40}
\providecommand{\natexlab}[1]{#1}
\providecommand{\url}[1]{\texttt{#1}}
\providecommand{\urlprefix}{URL }
\expandafter\ifx\csname urlstyle\endcsname\relax
  \providecommand{\doi}[1]{doi:\discretionary{}{}{}#1}\else
  \providecommand{\doi}{doi:\discretionary{}{}{}\begingroup
  \urlstyle{rm}\Url}\fi

\bibitem[{Bahdanau, Cho, and Bengio(2015)}]{attentionmechanism}
Bahdanau, D.; Cho, K.; and Bengio, Y. 2015.
\newblock Neural Machine Translation by Jointly Learning to Align and
  Translate.
\newblock In \emph{3rd International Conference on Learning Representations,
  {ICLR} 2015, San Diego, CA, USA, May 7-9, 2015, Conference Track
  Proceedings}.

\bibitem[{Bao et~al.(2019)Bao, Dong, Wei, Wang, Yang, Cui, Piao, and
  Zhou}]{uniencoding:bao}
Bao, H.; Dong, L.; Wei, F.; Wang, W.; Yang, N.; Cui, L.; Piao, S.; and Zhou, M.
  2019.
\newblock Inspecting Unification of Encoding and Matching with Transformer: A
  Case Study of Machine Reading Comprehension.
\newblock In \emph{Proceedings of the 2nd Workshop on Machine Reading for
  Question Answering}, 14--18. Association for Computational Linguistics.

\bibitem[{Bao et~al.(2020)Bao, Dong, Wei, Wang, Yang, Liu, Wang, Piao, Gao,
  Zhou, and Hon}]{unilmv2}
Bao, H.; Dong, L.; Wei, F.; Wang, W.; Yang, N.; Liu, X.; Wang, Y.; Piao, S.;
  Gao, J.; Zhou, M.; and Hon, H.-W. 2020.
\newblock {UniLMv2}: Pseudo-Masked Language Models for Unified Language Model
  Pre-Training.
\newblock \emph{arXiv preprint arXiv:2002.12804} .

\bibitem[{Bar-Haim et~al.(2006)Bar-Haim, Dagan, Dolan, Ferro, and
  Giampiccolo}]{rte2}
Bar-Haim, R.; Dagan, I.; Dolan, B.; Ferro, L.; and Giampiccolo, D. 2006.
\newblock The second {PASCAL} recognising textual entailment challenge.
\newblock In \emph{Proceedings of the Second {PASCAL} Challenges Workshop on
  Recognising Textual Entailment}.

\bibitem[{Bentivogli et~al.(2009)Bentivogli, Dagan, Dang, Giampiccolo, and
  Magnini}]{rte5}
Bentivogli, L.; Dagan, I.; Dang, H.~T.; Giampiccolo, D.; and Magnini, B. 2009.
\newblock The Fifth PASCAL Recognizing Textual Entailment Challenge.
\newblock In \emph{In Proc Text Analysis Conference}.

\bibitem[{Binder et~al.(2016)Binder, Montavon, Bach, M{\"{u}}ller, and
  Samek}]{lrp}
Binder, A.; Montavon, G.; Bach, S.; M{\"{u}}ller, K.; and Samek, W. 2016.
\newblock Layer-wise Relevance Propagation for Neural Networks with Local
  Renormalization Layers.
\newblock \emph{CoRR} abs/1604.00825.

\bibitem[{Brunner et~al.(2020)Brunner, Liu, Pascual, Richter, Ciaramita, and
  Wattenhofer}]{Brunner2020On}
Brunner, G.; Liu, Y.; Pascual, D.; Richter, O.; Ciaramita, M.; and Wattenhofer,
  R. 2020.
\newblock On Identifiability in Transformers.
\newblock In \emph{International Conference on Learning Representations}.

\bibitem[{Chen, Zheng, and Ji(2020)}]{chen2020Hierarchical}
Chen, H.; Zheng, G.; and Ji, Y. 2020.
\newblock Generating Hierarchical Explanations on Text Classification via
  Feature Interaction Detection.
\newblock In \emph{ACL}.

\bibitem[{Chi et~al.(2020{\natexlab{a}})Chi, Dong, Wei, Wang, Mao, and
  Huang}]{xnlg}
Chi, Z.; Dong, L.; Wei, F.; Wang, W.; Mao, X.; and Huang, H.
  2020{\natexlab{a}}.
\newblock Cross-Lingual Natural Language Generation via Pre-Training.
\newblock In \emph{The Thirty-Fourth {AAAI} Conference on Artificial
  Intelligence}, 7570--7577. {AAAI} Press.

\bibitem[{Chi et~al.(2020{\natexlab{b}})Chi, Dong, Wei, Yang, Singhal, Wang,
  Song, Mao, Huang, and Zhou}]{infoxlm}
Chi, Z.; Dong, L.; Wei, F.; Yang, N.; Singhal, S.; Wang, W.; Song, X.; Mao, X.;
  Huang, H.; and Zhou, M. 2020{\natexlab{b}}.
\newblock {InfoXLM}: An Information-Theoretic Framework for Cross-Lingual
  Language Model Pre-Training.
\newblock \emph{CoRR} abs/2007.07834.

\bibitem[{Clark et~al.(2019)Clark, Khandelwal, Levy, and
  Manning}]{what-bert-look}
Clark, K.; Khandelwal, U.; Levy, O.; and Manning, C.~D. 2019.
\newblock What Does {BERT} Look At? An Analysis of BERT's Attention.
\newblock \emph{CoRR} abs/1906.04341.

\bibitem[{Clark et~al.(2020)Clark, Luong, Le, and Manning}]{electra}
Clark, K.; Luong, M.-T.; Le, Q.~V.; and Manning, C.~D. 2020.
\newblock {ELECTRA}: Pre-training Text Encoders as Discriminators Rather Than
  Generators.
\newblock In \emph{ICLR}.

\bibitem[{Coenen et~al.(2019)Coenen, Reif, Yuan, Kim, Pearce, Vi{\'{e}}gas, and
  Wattenberg}]{geometry}
Coenen, A.; Reif, E.; Yuan, A.; Kim, B.; Pearce, A.; Vi{\'{e}}gas, F.~B.; and
  Wattenberg, M. 2019.
\newblock Visualizing and Measuring the Geometry of {BERT}.
\newblock \emph{CoRR} abs/1906.02715.

\bibitem[{Conneau et~al.(2020)Conneau, Khandelwal, Goyal, Chaudhary, Wenzek,
  Guzm{\'a}n, Grave, Ott, Zettlemoyer, and Stoyanov}]{xlmr}
Conneau, A.; Khandelwal, K.; Goyal, N.; Chaudhary, V.; Wenzek, G.; Guzm{\'a}n,
  F.; Grave, E.; Ott, M.; Zettlemoyer, L.; and Stoyanov, V. 2020.
\newblock Unsupervised Cross-lingual Representation Learning at Scale.
\newblock In \emph{Proceedings of the 58th Annual Meeting of the Association
  for Computational Linguistics}, 8440--8451. Association for Computational
  Linguistics.

\bibitem[{Dagan, Glickman, and Magnini(2006)}]{rte1}
Dagan, I.; Glickman, O.; and Magnini, B. 2006.
\newblock The PASCAL Recognising Textual Entailment Challenge.
\newblock In \emph{Proceedings of the First International Conference on Machine
  Learning Challenges: Evaluating Predictive Uncertainty Visual Object
  Classification, and Recognizing Textual Entailment}, MLCW'05, 177--190.
  Berlin, Heidelberg: Springer-Verlag.
\newblock ISBN 3-540-33427-0, 978-3-540-33427-9.

\bibitem[{Devlin et~al.(2019)Devlin, Chang, Lee, and Toutanova}]{bert}
Devlin, J.; Chang, M.-W.; Lee, K.; and Toutanova, K. 2019.
\newblock {BERT}: Pre-training of Deep Bidirectional {Transformers} for
  Language Understanding.
\newblock In \emph{Proceedings of the 2019 Conference of the North {A}merican
  Chapter of the Association for Computational Linguistics: Human Language
  Technologies, Volume 1}, 4171--4186. Minneapolis, Minnesota: Association for
  Computational Linguistics.

\bibitem[{Dolan and Brockett(2005)}]{mrpc2005}
Dolan, W.~B.; and Brockett, C. 2005.
\newblock Automatically constructing a corpus of sentential paraphrases.
\newblock In \emph{Proceedings of the Third International Workshop on
  Paraphrasing (IWP2005)}.

\bibitem[{Dong et~al.(2019)Dong, Yang, Wang, Wei, Liu, Wang, Gao, Zhou, and
  Hon}]{unilm}
Dong, L.; Yang, N.; Wang, W.; Wei, F.; Liu, X.; Wang, Y.; Gao, J.; Zhou, M.;
  and Hon, H.-W. 2019.
\newblock Unified Language Model Pre-training for Natural Language
  Understanding and Generation.
\newblock In \emph{33rd Conference on Neural Information Processing Systems
  (NeurIPS 2019)}.

\bibitem[{Ghaeini, Fern, and Tadepalli(2018)}]{lstm-attribution}
Ghaeini, R.; Fern, X.~Z.; and Tadepalli, P. 2018.
\newblock Interpreting Recurrent and Attention-Based Neural Models: a Case
  Study on Natural Language Inference.
\newblock \emph{CoRR} abs/1808.03894.

\bibitem[{Giampiccolo et~al.(2007)Giampiccolo, Magnini, Dagan, and
  Dolan}]{rte3}
Giampiccolo, D.; Magnini, B.; Dagan, I.; and Dolan, B. 2007.
\newblock The Third {PASCAL} Recognizing Textual Entailment Challenge.
\newblock In \emph{Proceedings of the {ACL}-{PASCAL} Workshop on Textual
  Entailment and Paraphrasing}, 1--9. Prague: Association for Computational
  Linguistics.

\bibitem[{Hewitt and Manning(2019)}]{hewitt-manning-2019-structural}
Hewitt, J.; and Manning, C.~D. 2019.
\newblock {A} Structural Probe for Finding Syntax in Word Representations.
\newblock In \emph{Proceedings of the 2019 Conference of the North {A}merican
  Chapter of the Association for Computational Linguistics: Human Language
  Technologies, Volume 1}, 4129--4138. Minneapolis, Minnesota: Association for
  Computational Linguistics.

\bibitem[{Hochreiter and Schmidhuber(1997)}]{lstm}
Hochreiter, S.; and Schmidhuber, J. 1997.
\newblock Long Short-Term Memory.
\newblock \emph{Neural Computation} 9: 1735--1780.
\newblock ISSN 0899-7667.

\bibitem[{Jain and Wallace(2019)}]{att-not}
Jain, S.; and Wallace, B.~C. 2019.
\newblock Attention is not Explanation.
\newblock \emph{CoRR} abs/1902.10186.

\bibitem[{Jin et~al.(2020)Jin, Wei, Du, Xue, and Ren}]{Jin2020Hierarchical}
Jin, X.; Wei, Z.; Du, J.; Xue, X.; and Ren, X. 2020.
\newblock Towards Hierarchical Importance Attribution: Explaining Compositional
  Semantics for Neural Sequence Models.
\newblock In \emph{ICLR}.

\bibitem[{Kovaleva et~al.(2019)Kovaleva, Romanov, Rogers, and
  Rumshisky}]{revealing-dark}
Kovaleva, O.; Romanov, A.; Rogers, A.; and Rumshisky, A. 2019.
\newblock Revealing the Dark Secrets of {BERT}.
\newblock In \emph{Proceedings of the 2019 Conference on Empirical Methods in
  Natural Language Processing and the 9th International Joint Conference on
  Natural Language Processing (EMNLP-IJCNLP)}, 4364--4373. Hong Kong, China:
  Association for Computational Linguistics.

\bibitem[{Liu et~al.(2019)Liu, Ott, Goyal, Du, Joshi, Chen, Levy, Lewis,
  Zettlemoyer, and Stoyanov}]{roberta}
Liu, Y.; Ott, M.; Goyal, N.; Du, J.; Joshi, M.; Chen, D.; Levy, O.; Lewis, M.;
  Zettlemoyer, L.; and Stoyanov, V. 2019.
\newblock {RoBERTa}: A Robustly Optimized {BERT} Pretraining Approach.
\newblock \emph{arXiv preprint arXiv:1907.11692} .

\bibitem[{Marecek and Rosa(2019)}]{Balustrades}
Marecek, D.; and Rosa, R. 2019.
\newblock From Balustrades to Pierre Vinken: Looking for Syntax in Transformer
  Self-Attentions.
\newblock \emph{CoRR} abs/1906.01958.

\bibitem[{Michel, Levy, and Neubig(2019)}]{are16headbetterthan1}
Michel, P.; Levy, O.; and Neubig, G. 2019.
\newblock Are Sixteen Heads Really Better than One?
\newblock \emph{CoRR} abs/1905.10650.

\bibitem[{Murdoch, Liu, and Yu(2018)}]{contextualdecompese}
Murdoch, W.~J.; Liu, P.~J.; and Yu, B. 2018.
\newblock Beyond Word Importance: Contextual Decomposition to Extract
  Interactions from LSTMs.
\newblock \emph{CoRR} abs/1801.05453.

\bibitem[{Murdoch and Szlam(2017)}]{celldecompose}
Murdoch, W.~J.; and Szlam, A. 2017.
\newblock Automatic Rule Extraction from Long Short Term Memory Networks.
\newblock \emph{CoRR} abs/1702.02540.

\bibitem[{Rosa and Marecek(2019)}]{inducing-trees}
Rosa, R.; and Marecek, D. 2019.
\newblock Inducing Syntactic Trees from {BERT} Representations.
\newblock \emph{CoRR} abs/1906.11511.

\bibitem[{Serrano and Smith(2019)}]{is-attention-interpret}
Serrano, S.; and Smith, N.~A. 2019.
\newblock Is Attention Interpretable?
\newblock In \emph{Proceedings of the 57th Annual Meeting of the Association
  for Computational Linguistics}, 2931--2951. Florence, Italy: Association for
  Computational Linguistics.

\bibitem[{Shrikumar, Greenside, and Kundaje(2017)}]{deeplift}
Shrikumar, A.; Greenside, P.; and Kundaje, A. 2017.
\newblock Learning Important Features Through Propagating Activation
  Differences.
\newblock \emph{CoRR} abs/1704.02685.

\bibitem[{Socher et~al.(2013)Socher, Perelygin, Wu, Chuang, Manning, Ng, and
  Potts}]{sst2013}
Socher, R.; Perelygin, A.; Wu, J.; Chuang, J.; Manning, C.~D.; Ng, A.; and
  Potts, C. 2013.
\newblock Recursive Deep Models for Semantic Compositionality Over a Sentiment
  Treebank.
\newblock In \emph{Proceedings of the 2013 Conference on Empirical Methods in
  Natural Language Processing}, 1631--1642. Seattle, Washington, USA:
  Association for Computational Linguistics.

\bibitem[{Sundararajan, Taly, and Yan(2017)}]{ig}
Sundararajan, M.; Taly, A.; and Yan, Q. 2017.
\newblock Axiomatic Attribution for Deep Networks.
\newblock \emph{CoRR} abs/1703.01365.

\bibitem[{Vaswani et~al.(2017)Vaswani, Shazeer, Parmar, Uszkoreit, Jones,
  Gomez, Kaiser, and Polosukhin}]{transformer}
Vaswani, A.; Shazeer, N.; Parmar, N.; Uszkoreit, J.; Jones, L.; Gomez, A.~N.;
  Kaiser, {\L}.; and Polosukhin, I. 2017.
\newblock Attention is All you Need.
\newblock In \emph{Advances in Neural Information Processing Systems 30},
  5998--6008. Curran Associates, Inc.

\bibitem[{Wallace et~al.(2019)Wallace, Feng, Kandpal, Gardner, and
  Singh}]{Wallace2019Triggers}
Wallace, E.; Feng, S.; Kandpal, N.; Gardner, M.; and Singh, S. 2019.
\newblock Universal Adversarial Triggers for Attacking and Analyzing {NLP}.
\newblock In \emph{Proceedings of the 2019 Conference on Empirical Methods in
  Natural Language Processing and the 9th International Joint Conference on
  Natural Language Processing (EMNLP-IJCNLP)}, 2153--2162. Hong Kong, China:
  Association for Computational Linguistics.

\bibitem[{Wang et~al.(2019)Wang, Singh, Michael, Hill, Levy, and
  Bowman}]{wang2018glue}
Wang, A.; Singh, A.; Michael, J.; Hill, F.; Levy, O.; and Bowman, S.~R. 2019.
\newblock {GLUE}: A Multi-Task Benchmark and Analysis Platform for Natural
  Language Understanding.
\newblock In \emph{International Conference on Learning Representations}.

\bibitem[{Wiegreffe and Pinter(2019)}]{att-not-not}
Wiegreffe, S.; and Pinter, Y. 2019.
\newblock Attention is not not Explanation.
\newblock In \emph{EMNLP-IJCNLP}, 11--20. Hong Kong, China: Association for
  Computational Linguistics.

\bibitem[{Williams, Nangia, and Bowman(2018)}]{mnli2017}
Williams, A.; Nangia, N.; and Bowman, S. 2018.
\newblock A Broad-Coverage Challenge Corpus for Sentence Understanding through
  Inference.
\newblock In \emph{Proceedings of the 2018 Conference of the North {A}merican
  Chapter of the Association for Computational Linguistics: Human Language
  Technologies, Volume 1 (Long Papers)}, 1112--1122. New Orleans, Louisiana:
  Association for Computational Linguistics.

\end{thebibliography}
